\documentclass[a4paper,fleqn]{cas-dc}

\usepackage[numbers]{natbib}

\usepackage{caption}

\usepackage{multirow}
\usepackage{booktabs}
\usepackage{algorithmicx}%
\usepackage[ruled,vlined,linesnumbered]{algorithm2e}
\usepackage{forloop} 
\usepackage[noend]{algpseudocode} 
\usepackage{xcolor}
\usepackage{subcaption}

\newcommand{\ciprian}[1]{{\footnotesize \color{orange} [{\bf Ciprian:} #1]}}
\definecolor{commentGreen}{rgb}{0,0.5,0.05}

\SetKwComment{Comment}{}{}

\def\tsc#1{\csdef{#1}{\textsc{\lowercase{#1}}\xspace}}
\tsc{WGM}
\tsc{QE}
\tsc{EP}
\tsc{PMS}
\tsc{BEC}
\tsc{DE}

\begin{document}
\let\WriteBookmarks\relax
\def\floatpagepagefraction{1}
\def\textpagefraction{.001}
\shorttitle{Multi-Dataset Cross-Domain Knowledge Distillation for Unified Medical Image Segmentation, Classification, and Detection}
\shortauthors{C.M. Ceausescu et~al.}

\title [mode = title]{Multi-Dataset Cross-Domain Knowledge Distillation for Unified Medical Image Segmentation, Classification, and Detection}

\tnotemark[1]

\tnotetext[1]{This research is supported by the project “Romanian Hub for Artificial Intelligence - HRIA”, Smart Growth, Digitization and Financial Instruments Program, 2021-2027, MySMIS no. 351416.}

\author[1]{Ciprian-Mihai Ceausescu}[type=editor,
                        auid=000,bioid=1,
                        orcid=0000-0002-0063-5095]
\cormark[1]
\ead{ciprian-mihai.ceausescu@drd.unibuc.ro}

\affiliation[1]{organization={Faculty of Mathematics and Computer Science, University of Bucharest},
                addressline={Str. Academiei, 14}, 
                city={Bucharest},
                postcode={010014}, 
                country={Romania}}

\author[1]{Ion-Marian Anghelina}[type=editor,
                        auid=000,bioid=1]

\author[1,2]{Dumitru-Bogdan Alexe}[type=editor,
                        auid=000,bioid=1,
                        orcid=0009-0004-9732-8770]

\affiliation[2]{organization={Gheorghe Mihoc-Caius Iacob Institute of Mathematical Statistics and Applied Mathematics of the Romanian Academy, Romania},
                addressline={Calea 13 Septembrie, 13}, 
                city={Bucharest},
                postcode={050711}, 
                country={Romania}}

\cortext[cor1]{Corresponding author}

\begin{abstract}
We propose a unified cross-domain transfer learning framework that leverages knowledge from multiple heterogeneous medical imaging datasets to improve performance across segmentation, classification, and object detection tasks. Our approach employs a teacher–student paradigm in which a joint teacher model aggregates domain-invariant representations learned from diverse source datasets, while a task-specific student model is trained via multi-level knowledge distillation. Originally developed for medical image segmentation, the framework is extended to support image-level classification and object-level detection, enabling a general multi-task formulation for medical image analysis. We evaluate our method on a broad suite of datasets, including six segmentation benchmarks, BrainMetShare, ISLES, BraTS (MRI) and Lung MSD, LiTS, KiTS (CT), as well as multiple classification datasets for pulmonary disease and dementia, and detection datasets with native bounding-box annotations. Across all tasks and modalities, the proposed approach yields consistent improvements over strong dataset-specific and multi-head baselines, demonstrating enhanced robustness to distributional shifts and superior generalization. These findings highlight the potential of multi-dataset knowledge distillation as a scalable and task-agnostic approach for enhancing segmentation, classification, and object detection performance across heterogeneous medical imaging domains.
\end{abstract}


\maketitle

\begin{abstract}
We propose a unified cross-domain transfer learning framework that leverages knowledge from multiple heterogeneous medical imaging datasets to improve performance across segmentation, classification, and object detection tasks. Our approach employs a multi-dataset teacher–student paradigm, in which multiple source and target teachers are first trained on individual datasets and subsequently fused into a single joint teacher that aggregates domain-invariant representations learned across sources. A task-specific student model is then trained through multi-level knowledge distillation from this joint teacher. Originally developed for medical image segmentation, the framework is extended in this journal version to support image-level classification and object-level detection, thereby enabling a general multi-task formulation for medical image analysis. We evaluate our method on a broad suite of datasets, including six segmentation benchmarks, BrainMetShare, ISLES, BraTS (MRI), and Lung MSD, LiTS, KiTS (CT), as well as multiple classification datasets for pulmonary disease and dementia, and detection datasets with native bounding-box annotations. Across all tasks and modalities, the proposed approach yields consistent improvements over strong dataset-specific and multi-head baselines, demonstrating enhanced robustness to distributional shifts and superior generalization. These findings highlight the potential of multi-dataset knowledge distillation as a scalable and task-agnostic approach for enhancing segmentation, classification, and object detection performance across heterogeneous medical imaging domains.

\end{abstract}

\section{Introduction}
\label{main}
Deep learning has become central to modern medical image analysis, enabling automated interpretation of MRI, CT, and X-ray scans for tasks such as tumor delineation, disease classification, and lesion detection. Accurate segmentation of anatomical and pathological structures is essential for clinical decision-making in oncology, neurology, and pulmonary medicine. Over the past decade, architectures based on convolutional networks and transformers~\cite{cao2022eccvw, chen2021transunet, chen2020icip, ronneberger2015u,wenjian-2024} have achieved impressive progress, particularly when trained in a supervised fashion on large, well-curated datasets. However, in medical imaging, acquiring large quantities of expert-annotated data remains extremely challenging due to annotation costs, limited patient access, and variations in imaging protocols across clinical centers.

To address data scarcity, prior work has explored augmentation strategies and generative modeling to synthetically expand training sets. Early work with UNet~\cite{ronneberger2015u} showed that heavy elastic deformations can partially compensate for limited labels. More sophisticated approaches employ generative adversarial networks~\cite{mok2019springer,sandfort2019data}, variational autoencoders~\cite{ding2021modeling}, and diffusion models~\cite{chen2023berdiff,zhang2024diffboost} to synthesize realistic medical images. While these techniques improve the amount of training data, generative methods often fail to capture the fine-grained anatomical details and inter-dataset variability necessary for robust generalization.

Transfer learning offers an alternative by adapting models pre-trained on large-scale natural image datasets to medical tasks~\cite{hosseinzadeh2021systematic, kim2022transfer, shin2016deep}. Fine-tuned ImageNet models~\cite{xie2018pre} and domain-adapted anomaly detection networks~\cite{ceausescu2024coreset} have demonstrated promising performance in medical segmentation and classification. Nevertheless, traditional transfer learning is limited by its reliance on single-source pre-training, which is insufficient to overcome the wide range of domain shifts encountered in medical imaging.

A promising direction involves knowledge distillation, where a compact student network learns from a larger teacher model. Yet most distillation approaches rely on a single teacher, assume homogeneous data distributions, or focus exclusively on one task (e.g., segmentation). In practice, medical imaging workflows encompass a diverse set of tasks, including segmentation, classification, and detection, each influenced by dataset-specific nuances such as modality, scanner type, pathology appearance, and annotation granularity. A robust distillation framework must, therefore, integrate heterogeneous sources of knowledge, align cross-domain feature representations, and support multiple downstream tasks.

In this work, we introduce a unified multi-dataset cross-domain knowledge distillation framework designed to improve generalization across diverse medical imaging tasks, including segmentation, classification, and object detection. Our approach trains multiple teacher models, each associated with a different dataset or domain, using a domain-adversarial objective that encourages alignment between the source and target feature distributions. We then construct a joint teacher by fusing multi-level encoder and bottleneck features via cross-attention mechanisms, enabling the aggregation of complementary representations learned from heterogeneous data sources. A compact student model is trained to inherit these multi-source representations through a curriculum-driven distillation strategy that combines segmentation, contrastive, feature-alignment, and similarity-based objectives, depending on the target task.


We evaluate our method on a large and heterogeneous collection of medical datasets. For segmentation, we use six MRI and CT datasets spanning multiple anatomical regions: BrainMetShare~\cite{grovik2020deep}, ISLES 2022~\cite{hernandez2022isles,de2024robust}, BraTS~\cite{menze2014multimodal}, Lung MSD~\cite{antonelli2022medical}, LiTS~\cite{bilic2023liver}, and KiTS~\cite{heller2019kits19}. To assess classification, we use pulmonary disease and dementia datasets: COVIDx-CXR~\cite{pavlova2022covidxcxr3largescaleopensource}, RT-PCR-COVID19~\cite{cohen2020covid19imagedatacollection}, 
COVID-QU-Ex~\cite{chowdhury2020screening, degerli2021infectionmap,rahman2021imageenhancement, TAHIR2021105002, tahir2021covidquex}, OASIS MRI ~\cite{marcus2007oasis}, and ADNI~\cite{mueller2005adni}. For object detection, we employ datasets with native bounding-box annotations including Lung Cancer CT \& PET-CT~\cite{lung_cancer_ct_petct}, LungCT~\cite{lungct_roboflow} and DeepLesion~\cite{yan2018deeplesion}. This broad experimental setup allows us to analyze the robustness of the proposed framework across different tasks, modalities, anatomical regions, and label granularities.

This journal article represents a substantial extension of our preliminary work~\cite{Ceausescu25kes}. While the conference paper focused exclusively on segmentation and evaluated the method on six datasets, the present version:
(i) extends the framework to classification and object detection;
(ii) introduces additional datasets across three imaging modalities;
(iii) provides a deeper analysis of cross-domain fusion strategies and curriculum-based distillation; (iv) delivers a significantly expanded experimental study, including ablations, per-dataset evaluations, and cross-architecture transfer experiments.

Our main contributions are: (1) a unified multi-dataset distillation framework capable of supporting segmentation, classification, and object detection; (2) a novel joint-teacher architecture that fuses multi-level features from multiple source teachers via cross-attention, enabling robust cross-domain representation transfer; (3) a curriculum-based distillation objective combining segmentation, contrastive, alignment, and similarity losses; (4) a comprehensive evaluation across three tasks, three modalities (MRI, CT, X-ray), and numerous datasets, showing consistent improvements over dataset-specific baselines and multi-head multi-dataset models.

\section{Related Work}

Medical image analysis spans a wide range of tasks, including segmentation, classification, and object detection, all of which are highly sensitive to variations in imaging modality, acquisition protocol, scanner hardware, and patient population. As a consequence, models trained on a single dataset often exhibit poor generalization when deployed in new clinical environments, motivating extensive research on transfer learning, domain adaptation, multi-task learning, and knowledge distillation~\cite{glocker2019machine,guan2021domain,zech2018variable}.

\medskip
\noindent \textit{Transfer learning and domain adaptation.}
A common strategy to mitigate data scarcity is fine-tuning models pre-trained on large-scale natural image datasets such as ImageNet~\cite{deng2009imagenet} or ADE20k~\cite{zhou2017scene}. Numerous studies~\cite{hosseinzadeh2021systematic,patrascu2024ssspr,shin2016deep,xie2018pre} have shown that transfer learning provides strong baselines for medical image classification and segmentation, even under significant domain shifts. However, fine-tuning inherently follows a single source--target paradigm and does not explicitly address dataset bias or the heterogeneity arising from multiple complementary medical datasets~\cite{zech2018variable}.

Domain adaptation methods aim to reduce discrepancies between source and target distributions, enabling knowledge transfer across institutions or modalities. Adversarial approaches such as DANN~\cite{ganin2016domain} and its extensions~\cite{omidi2024unsupervised} align feature representations using domain discriminators, while other methods match higher-order statistics (e.g., DeepCORAL~\cite{sun2016deep}) or rely on image translation models such as CycleGAN~\cite{zhu2017unpaired}. Although effective, most domain adaptation techniques assume a single source domain and do not exploit information distributed across multiple heterogeneous datasets.

\medskip
\noindent \textit{Medical image segmentation.}
Segmentation is a cornerstone task in medical image analysis, underpinning applications such as tumor delineation, organ quantification, and treatment planning. Convolutional encoder–decoder architectures, most notably U-Net and its variants~\cite{cciccek20163d,ronneberger2015u,zhou2018unet++}, have become the de facto standard due to their ability to capture fine-grained spatial detail while maintaining global context. Subsequent works incorporate attention mechanisms~\cite{oktay2018attention}, multi-scale feature aggregation~\cite{ibtehaz2020multiresunet}, and transformer-based components~\cite{chen2021transunet,hatamizadeh2022unetr} to further improve segmentation accuracy. Despite these advances, segmentation models are highly sensitive to domain shifts caused by differences in imaging protocols, scanners, and patient populations. Most existing approaches are trained and evaluated on a single dataset, limiting their robustness and generalization when deployed across institutions or modalities. This motivates learning strategies that can leverage complementary information from multiple datasets while preserving task-specific precision.

\medskip
\noindent \textit{Classification and detection in medical imaging.}
Deep learning has significantly advanced medical image classification and detection, with convolutional neural networks (CNNs)~\cite{khalifa2024,Mienye2025,shen2017deep} remaining a dominant paradigm due to their strong inductive bias for local structure. Extensive surveys and empirical studies~\cite{esteva2019guide,glocker2019machine, litjens2017survey,suzuki2017overview} report consistent gains in tumor classification, lesion detection, and organ localization, while highlighting persistent challenges related to generalization, domain shift, and annotation scarcity. More recently, Vision Transformers (ViTs) and attention-based architectures~\cite{aburass2025, azad2024} have gained traction for their ability to capture long-range dependencies, often achieving competitive performance when adapted to medical imaging. Complementary directions include self-supervised and semi-supervised learning~\cite{huang2023selfsupervised}, generative models for data augmentation and privacy preservation~\cite{khosravi2025}, and explainable AI~\cite{houssein2025}. Despite these advances, most approaches remain confined to single-dataset or single-task settings.

\medskip
\noindent \textit{Multi-dataset and multi-task learning.}
To better reflect clinical heterogeneity, recent work increasingly explores multi-dataset and multi-task learning paradigms. Multi-dataset learning leverages data from different institutions, scanners, or modalities to improve robustness and reduce sensitivity to dataset-specific biases~\cite{glocker2019machine,guan2021domain,zech2018variable}. In parallel, multi-task learning (MTL) trains shared representations across related tasks such as classification, segmentation, and detection, improving data efficiency and regularization~\cite{ruder2017overview}. Systematic reviews~\cite{zhao2023multitask} show that MTL architectures with shared encoders or attention-based routing often outperform isolated single-task models. Recent efforts further integrate transformers and cross-task attention to model task interactions~\cite{bui2024multiscalefeature}, while initiatives such as MedIMeta~\cite{woerner2025} provide unified benchmarks for multi-domain, multi-task evaluation.

\medskip
\noindent \textit{Knowledge distillation.}
Knowledge distillation (KD)~\cite{cheng2020explaining, cho2019efficacy, hinton2015distilling} provides a principled mechanism for transferring information from a high-capacity teacher to a compact or task-specific student. Beyond model compression, KD has been applied to cross-modality transfer~\cite{jiang2021unpaired} and unsupervised domain adaptation~\cite{yao2022novel}. In medical imaging, however, most distillation-based approaches rely on a single teacher trained on a single dataset and are typically designed for a single downstream task, limiting their ability to exploit complementary knowledge distributed across heterogeneous datasets and clinical domains.

Recent work has emphasized the importance of \emph{multi-level} feature supervision in distillation. In particular, Multi-Level Feature Distillation (MLFD)~\cite{iordache2024mlfd} shows that distilling representations from multiple intermediate layers of a single teacher provides a richer supervisory signal than output-level distillation alone. By aligning features across different abstraction levels, MLFD captures both low-level structural cues and high-level semantic information, leading to improved robustness and generalization on natural image benchmarks. While highly relevant conceptually, MLFD is evaluated exclusively on natural image benchmarks and does not address the domain shift, dataset heterogeneity, or task diversity inherent to medical imaging. In particular, it focuses on image-level recognition tasks and does not consider dense prediction problems such as medical image segmentation or object detection, which pose fundamentally different challenges in terms of spatial alignment and supervision.

Several subsequent works further explore structured feature distillation across layers, architectures, or pretraining regimes, reporting improved generalization under distribution shifts~\cite{liu2020,wang2024,zhang2025}. Despite these advances, the potential of multi-level distillation remains largely unexplored in medical imaging, particularly in settings involving multiple datasets, imaging modalities, and tasks.

\medskip
\noindent \textit{Positioning of our work.}
Our work builds on these insights by introducing a unified cross-domain teacher-student framework tailored to medical imaging. We focus on aggregating knowledge across multiple heterogeneous medical datasets and extending distillation beyond segmentation to classification and object detection. By constructing a joint teacher that fuses multi-level representations via cross-attention and distilling this knowledge into task-specific students, our framework addresses dataset heterogeneity, domain shift, and task diversity within a single, scalable formulation.

\begin{figure*}[!ht]
    \centering
    \includegraphics[width=0.75\linewidth]{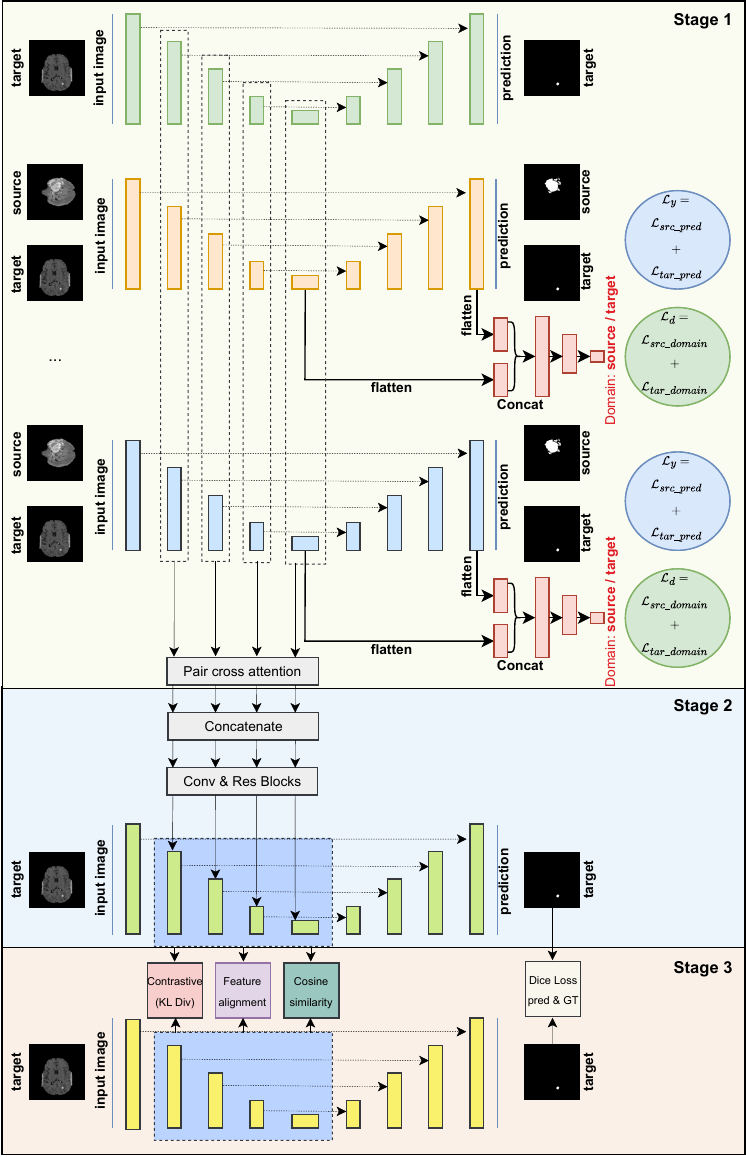}
    \caption{\emph{Overview of our pipeline.} 
\textbf{Stage 1:} Teacher models are trained on both the target and source tasks. The target dataset $ \mathbf{D}^t $ is incorporated into the training of source teacher models to align the feature distributions between the source and target domains. 
\textbf{Stage 2:} A joint teacher model is constructed by integrating features from the encoder and bottleneck of the target and source teachers at corresponding levels of abstraction. All encoder and bottleneck features are fused while remaining frozen during the fusion process, ensuring knowledge retention. The decoder is trained from scratch on the target dataset $\mathbf{D}^t$ to optimize segmentation performance.
\textbf{Stage 3:} Knowledge is distilled from the joint teacher to the student model using examples exclusively from the target dataset  $\mathbf{D}^t$.}
    \label{fig:teacher-fusion}
\end{figure*}

\section{Methods}

Our framework receives as input: (i) a collection of datasets 
$\mathcal{D}=\{\mathbf{D}^t,\mathbf{D}^{s_1},\ldots,\mathbf{D}^{s_m}\}$; 
(ii) a matching set of teacher networks 
$\mathcal{T}=\{\mathcal{T}^t,\mathcal{T}^{s_1},\ldots,\mathcal{T}^{s_m}\}$; (iii) a student network $\mathcal{S}^t$ associated with the target dataset $\mathbf{D}^t$. Each dataset $\mathbf{D}^{d} = \{(I_i^{d}, y_i^{d})\}_{i=1}^{n_d}$ contains images $I_i^{d}$ and corresponding annotations $y_i^{d}$, which can include segmentation masks (for segmentation), class labels (for classification), or bounding-box annotations (for detection). The target dataset is denoted by $d = t$, while $d = s_k$ with $k \in \{1,\dots,m\}$ indexes the source datasets.

Our goal is to leverage the heterogeneous source datasets $\{\mathbf{D}^{s_k}\}_{k=1}^m$ to train a set of teacher models $\{\mathcal{T}^{s_k}\}_{k=1}^m$ and a target teacher $\mathcal{T}^t$, then aggregate their knowledge into a \emph{joint teacher} $\mathcal{T}_*$, and finally distill this multi-source knowledge into a compact, task-specific student $\mathcal{S}^t$. The proposed framework consists of three sequential stages: 

\begin{enumerate}
    \item \textbf{Stage 1:} Domain-adversarial training of source and target teachers. 
    \item \textbf{Stage 2:} Construction of a joint teacher $\mathcal{T}_*$ through cross-attention fusion of multi-level features.
    \item \textbf{Stage 3:} Multi-level distillation from $\mathcal{T}_*$ to the student $\mathcal{S}^t$ using a curriculum-driven objective. 
\end{enumerate}

We first describe the proposed framework in the context of medical image segmentation, which represents the primary application scenario of our work. The overall training procedure is summarized in Algorithm~\ref{alg_CDTS}. With minimal modifications to the task-specific components, the same pipeline can be applied to image classification and object detection, as discussed later in this section. For clarity, we detail the loss formulations for the segmentation setting, where the supervision labels $y_i^{d}$ correspond to segmentation masks $M_i^{d}$, and subsequently outline the adaptations required for classification and detection.

\begin{figure*}[t!]
    \centering
    \includegraphics[scale=1.35]{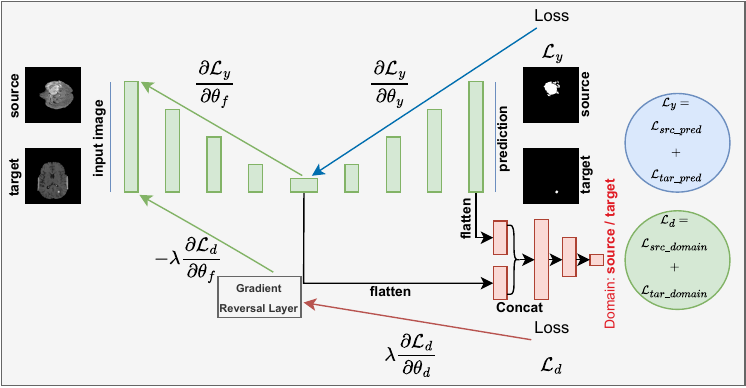}
    \caption{\emph{Teacher model $ \mathcal{T}^{s_k} $ is trained using a domain adaptation strategy.} The losses $ \mathcal{L}_{y} $ and $ \mathcal{L}_{d} $ are computed to update the model parameters, thus enabling the encoder to learn domain-invariant features.}    
    \label{fig:domain-adaptation-1}
\end{figure*}

\begin{algorithm*}[!t]
\small
\caption{Cross-domain teacher-student framework for medical image segmentation}
\label{alg_CDTS}
\KwIn{
Target dataset $\mathbf{D}^{t}=\{(I^{t}_i,M^{t}_i)\}_{i=1}^{n_t}$;
source datasets $\{\mathbf{D}^{s_k}=\{(I^{s_k}_i,M^{s_k}_i)\}_{i=1}^{n_{s_k}}\}_{k=1}^{m}$;
target teacher $\mathcal{T}^{t}$; source teachers $\{\mathcal{T}^{s_k}\}_{k=1}^m$;
domain discriminator $\mathcal{D}$ (attached to each $\mathcal{T}^{s_k}$);
student $\mathcal{S}^{t}$.\\
Learning rates $\eta_{t},\eta_{s},\eta_{*},\eta_{S}$;  batch sizes $b_1,b_2$; adaptation flag $r\in\{0,1\}$;
adversarial schedule $\lambda(\rho)$; curriculum schedule $cf(e)$.
}
\KwOut{Trained student parameters $\theta_{\mathcal{S}^{t}}$.}

{\textcolor{commentGreen}{$\lhd$ Stage 1: Train teachers (target + source with domain adaptation)}}

Initialize $\theta_{\mathcal{T}^t}$ ;
{\textcolor{commentGreen}{$\lhd$ initialize target teacher parameters}}

\For({\textcolor{commentGreen}{$\lhd$ iterate over training epochs}}){$e = 1$ \KwTo $E_t$}{
    \ForEach({\textcolor{commentGreen}{$\lhd$ iterate over target mini-batches}}){mini-batch $\mathcal{B}^t=\{(I^t,M^t)\}$ from $\mathbf{D}^t$}{
        $\hat{M}^t \leftarrow \mathcal{T}^t(I^t)$ ;
        {\textcolor{commentGreen}{$\lhd$ forward pass through target teacher}}

        $\mathcal{L}^t \leftarrow \mathrm{Dice}(\hat{M}^t,M^t)$ ;
        {\textcolor{commentGreen}{$\lhd$ compute segmentation loss on target data}}

        $\theta_{\mathcal{T}^t} \leftarrow \theta_{\mathcal{T}^t} - \eta_t \nabla_{\theta_{\mathcal{T}^t}}\mathcal{L}^t$ ;
        {\textcolor{commentGreen}{$\lhd$ update target teacher parameters}}
    }
}

\ForEach({\textcolor{commentGreen}{$\lhd$ iterate over source datasets}}){$k \in \{1,\ldots,m\}$}{ 

    Initialize $\theta_{\mathcal{T}^{s_k}} = \{\theta_{f}^{s_k},\theta_{y}^{s_k},\theta_{d}^{s_k}\}$ ;
    {\textcolor{commentGreen}{$\lhd$ initialize source teacher (encoder, decoder, discriminator)}}

    \For({\textcolor{commentGreen}{$\lhd$ iterate over training epochs}}){$e = 1$ \KwTo $E_s$}{
        \ForEach({\textcolor{commentGreen}{$\lhd$ paired source--target mini-batches}}){$\mathcal{B}^{s_k}, \mathcal{B}^{t}$}{

            $F^{s} \leftarrow f_{\mathcal{T}^{s_k}}(I^{s_k}), \;
             F^{t} \leftarrow f_{\mathcal{T}^{s_k}}(I^{t})$ ;
            {\textcolor{commentGreen}{$\lhd$ extract encoder features for source and target samples}}

            $x^{s} \leftarrow \mathcal{D}(F^{s}), \;
             x^{t} \leftarrow \mathcal{D}(F^{t})$ ;
            {\textcolor{commentGreen}{$\lhd$ predict domain labels via discriminator}}

            $\hat{M}^{s} \leftarrow \mathcal{T}^{s_k}(I^{s_k}), \;
             \hat{M}^{t} \leftarrow \mathcal{T}^{s_k}(I^{t})$ ;
            {\textcolor{commentGreen}{$\lhd$ segmentation predictions for source and target}}

            $\mathcal{L}^{s_k}_{y} \leftarrow 
                \sum_{i=1}^{b_1}\mathrm{Dice}(\hat{M}^{s}_i,M^{s}_i)
                + r\sum_{j=1}^{b_2}\mathrm{Dice}(\hat{M}^{t}_j,M^{t}_j)$ ;
            {\textcolor{commentGreen}{$\lhd$ task loss with optional supervised target adaptation}}

            $\mathcal{L}^{s_k}_{d} \leftarrow 
                \sum_{i=1}^{b_1}\mathrm{BCE}(x^{s}_i,1)
                + \sum_{j=1}^{b_2}\mathrm{BCE}(x^{t}_j,0)$ ;
            {\textcolor{commentGreen}{$\lhd$ domain-adversarial loss}}

            $\theta_{y}^{s_k} \leftarrow \theta_{y}^{s_k} - \eta_s \nabla_{\theta_{y}^{s_k}}\mathcal{L}^{s_k}_{y}$ ;
            {\textcolor{commentGreen}{$\lhd$ update segmentation head}}

            $\theta_{d}^{s_k} \leftarrow \theta_{d}^{s_k} - \eta_s \lambda(\rho)\nabla_{\theta_{d}^{s_k}}\mathcal{L}^{s_k}_{d}$ ;
            {\textcolor{commentGreen}{$\lhd$ update discriminator}}

            $\theta_{f}^{s_k} \leftarrow \theta_{f}^{s_k}
            - \eta_s\big(\nabla_{\theta_{f}^{s_k}}\mathcal{L}^{s_k}_{y}
            - \lambda(\rho)\nabla_{\theta_{f}^{s_k}}\mathcal{L}^{s_k}_{d}\big)$ ;
            {\textcolor{commentGreen}{$\lhd$ adversarial encoder update via gradient reversal}}
        }
    }
}

\BlankLine
{\textcolor{commentGreen}{$\lhd$ Stage 2: Build and train the joint teacher via multi-level fusion}}

$\mathcal{T}_* \leftarrow \textsc{BuildJointTeacher}\!\big(\mathcal{T}^{t}, \{\mathcal{T}^{s_k}\}_{k=1}^{m}\big)$ ;
{\textcolor{commentGreen}{$\lhd$ fuse multi-level teacher features using cross-attention}}

$\textsc{Freeze}(\mathcal{T}^{t}.f, \mathcal{T}^{t}.b,\{\mathcal{T}^{s_k}.f,\mathcal{T}^{s_k}.b\}_{k=1}^{m})$ ;
{\textcolor{commentGreen}{$\lhd$ freeze all teacher encoders and bottlenecks}}

Initialize $\theta_{g_*}$ ;
{\textcolor{commentGreen}{$\lhd$ initialize task-specific head for the joint teacher}}

\For({\textcolor{commentGreen}{$\lhd$ iterate over training epochs}}){$e = 1$ \KwTo $E_*$}{
    \ForEach({\textcolor{commentGreen}{$\lhd$ iterate over target mini-batches}}){$\mathcal{B}^{t}$}{
        $\{f_*^{(\ell)}\} \leftarrow \mathcal{T}_*.\textsc{Backbone}(I^{t})$ ;
        {\textcolor{commentGreen}{$\lhd$ extract fused multi-level features}}

        $\hat{M}_* \leftarrow g_*(\{f_*^{(\ell)}\})$ ;
        {\textcolor{commentGreen}{$\lhd$ predict segmentation using joint teacher head}}

        $\mathcal{L}_* \leftarrow \mathrm{Dice}(\hat{M}_*,M^{t})$ ;
        {\textcolor{commentGreen}{$\lhd$ compute joint teacher loss}}

        $\theta_{g_*} \leftarrow \theta_{g_*} - \eta_* \nabla_{\theta_{g_*}}\mathcal{L}_*$ ;
        {\textcolor{commentGreen}{$\lhd$ update joint teacher head only}}
    }
}

\BlankLine
{\textcolor{commentGreen}{$\lhd$ Stage 3: Distill multi-level knowledge from the joint teacher to the student}}

Initialize $\theta_{\mathcal{S}^{t}}$ ;
{\textcolor{commentGreen}{$\lhd$ initialize student network}}

\For({\textcolor{commentGreen}{$\lhd$ iterate over training epochs}}){$e = 1$ \KwTo $E_S$}{
    \ForEach({\textcolor{commentGreen}{$\lhd$ iterate over target mini-batches}}){$\mathcal{B}^{t}$}{

        $(\hat{M}_S,\{f_S^{(\ell)}\}) \leftarrow \mathcal{S}^{t}(I^{t})$ ;
        {\textcolor{commentGreen}{$\lhd$ student forward pass and feature extraction}}

        $(\hat{M}_T,\{f_T^{(\ell)}\}) \leftarrow \mathcal{T}_*(I^{t})$ ;
        {\textcolor{commentGreen}{$\lhd$ joint teacher forward pass (frozen)}}

        $\mathcal{L}_{\text{task}} \leftarrow \mathrm{DiceBCE}(\hat{M}_S,M^{t})$ ;
        {\textcolor{commentGreen}{$\lhd$ compute supervised student loss}}

        $\mathcal{L}_{\text{FAlgn}} \leftarrow \sum_{\ell}\|f_S^{(\ell)}-f_T^{(\ell)}\|_2^2$ ;
        {\textcolor{commentGreen}{$\lhd$ feature-level alignment loss}}

        $\mathcal{L}_{\text{CosSim}} \leftarrow \sum_{\ell}\big(1-\cos(f_S^{(\ell)},f_T^{(\ell)})\big)$ ;
        {\textcolor{commentGreen}{$\lhd$ representation-level cosine alignment}}

        $\mathcal{L}_{\text{Con}} \leftarrow \text{batch-contrastive}(\{f_S^{(\ell)}\},\{f_T^{(\ell)}\})$ ;
        {\textcolor{commentGreen}{$\lhd$ batch-wise relational distillation}}

        $\mathcal{L}_{\text{total}} \leftarrow \alpha \mathcal{L}_{\text{task}}
        + cf(e)\big(\beta \mathcal{L}_{\text{Con}} + \gamma \mathcal{L}_{\text{FAlgn}} + \delta \mathcal{L}_{\text{CosSim}}\big)$ ;
        {\textcolor{commentGreen}{$\lhd$ curriculum-weighted total loss}}

        $\theta_{\mathcal{S}^{t}} \leftarrow \theta_{\mathcal{S}^{t}} - \eta_S \nabla_{\theta_{\mathcal{S}^{t}}}\mathcal{L}_{\text{total}}$ ;
        {\textcolor{commentGreen}{$\lhd$ update student parameters}}
    }
}
\Return $\theta_{\mathcal{S}^{t}}$ \;
\end{algorithm*}

\vspace{2mm}
\subsection{Stage 1: Teacher models with domain adversarial alignment}

\vspace{2mm}
\noindent{\bf Target teacher.}
The target teacher $\mathcal{T}^t$ is trained exclusively on the target dataset $\mathbf{D}^t$ using a segmentation loss (Algorithm~\ref{alg_CDTS}, stage 1, lines~1--7). In our experiments, we employ the Dice loss, defined for a mini-batch of size $b$ as:
\begin{equation}
    \mathcal{L}^t = \sum_{i=1}^{b} \text{Dice}(\hat{M}_i^t, M_i^t),
\label{eq:loss_t}
\end{equation}
where $\hat{M}_i^t$ and $M_i^t$ denote the predicted and ground-truth masks for the $i$-th target image, respectively.\\

\noindent{\bf Source teachers with domain adaptation.}
Each source teacher $\mathcal{T}^{s_k}$, with $k\in\{1,\ldots,m\}$, is trained on its own source dataset $\mathbf{D}^{s_k}$ and on the target dataset $\mathbf{D}^t$ (Algorithm~\ref{alg_CDTS}, stage 1, lines~8--19) to encourage domain-invariant representations (Figure~\ref{fig:domain-adaptation-1}). Following the domain-adversarial strategy of~\cite{omidi2024unsupervised}, each teacher is augmented with a domain discriminator that operates on intermediate features.

Let $b_1$ and $b_2$ be the number of source and target images in a mini-batch, respectively. The loss for teacher $\mathcal{T}^{s_k}$ is:
\begin{equation}
\begin{aligned}
    \mathcal{L}^{s_k} &= \mathcal{L}^{s_k}_y + \mathcal{L}^{s_k}_d, \\[4pt]
    \mathcal{L}^{s_k}_y &= 
        \sum_{i=1}^{b_1} 
        \text{Dice}(\hat{M}_i^{s_k}, M_i^{s_k})
        + r \cdot 
        \sum_{i=1}^{b_2} 
        \text{Dice}(\hat{M}_i^{t}, M_i^{t}), \\[4pt]
    \mathcal{L}^{s_k}_d &= 
        \sum_{i=1}^{b_1} 
        \text{BCE}(x_i^{s_k}, d_i^{s_k})
        + 
        \sum_{j=1}^{b_2} 
        \text{BCE}(x_j^{t}, d_j^{t}),
\end{aligned}
\label{eq:loss_divided_1}
\end{equation}
where $\mathcal{L}^{s_k}_y$ is the segmentation loss (or the task-specific loss in classification/detection settings), while $\mathcal{L}^{s_k}_d$ is the discriminator loss that encourages the encoder to produce domain-invariant features by making source and target samples indistinguishable to the discriminator. $\text{BCE}(\cdot)$ denotes the binary cross-entropy loss. The parameter $r \in \{0,1\}$ controls whether labeled target data are used during training: $r=0$ corresponds to unsupervised domain adaptation, whereas $r=1$ leverages labeled target examples to directly optimize segmentation performance. 
In our experimental setup, labeled target annotations are available and are therefore exploited by setting $r=1$. 
This choice allows the source teachers to better align their representations to the target task and to produce stronger supervision signals for the subsequent fusion and distillation stages. 
Nevertheless, to assess the robustness of the framework when target labels are unavailable, we also analyze the the unsupervised variant ($r=0$) in a dedicated ablation study (Section~\ref{sec:ablations}), where we show that the proposed method remains effective even without target supervision, while supervised adaptation primarily improves boundary accuracy.
%
In this formulation, $d_i^{s_k} = 1$ and $d_j^{t} = 0$ represent the ground-truth domain labels for source and target samples, respectively.  The terms $x_i^{s_k}$ and $x_j^{t}$ are the corresponding discriminator predictions,  obtained by applying the domain discriminator $\mathcal{D}$ to the intermediate 
encoder feature maps of each sample:
\begin{equation}
\begin{aligned}
x_i^{s_k} = \mathcal{D}\!\left(f_{\mathcal{T}^{s_k}}(I_i^{s_k})\right), 
\qquad
x_j^{t} = \mathcal{D}\!\left(f_{\mathcal{T}^{s_k}}(I_j^{t})\right),
\end{aligned}
\label{eq:xs}
\end{equation}
where $f_{\mathcal{T}^{s_k}}(\cdot)$ denotes the encoder of the teacher 
$\mathcal{T}^{s_k}$.

Each teacher $\mathcal{T}^{s_k}$ is decomposed into an encoder, a decoder, and a domain discriminator with parameters $\theta_{\textbf{T}^{s_k},f}$, $\theta_{\textbf{T}^{s_k},y}$, and $\theta_{\textbf{T}^{s_k},d}$, respectively. The update rules are:
\begin{equation}
    \theta_{\textbf{T}^{s_k},y} \leftarrow \theta_{\textbf{T}^{s_k},y}
    - \eta_{T^{s_k}} \frac{\partial \mathcal{L}_y^{s_k}}{\partial \theta_{\textbf{T}^{s_k},y}},
\label{eq:update-decoder}
\end{equation}
\begin{equation}
    \theta_{\textbf{T}^{s_k},d} \leftarrow \theta_{\textbf{T}^{s_k},d}
    - \eta_{T^{s_k}} \lambda \frac{\partial \mathcal{L}_d^{s_k}}{\partial \theta_{\textbf{T}^{s_k},d}},
\label{eq:update-discriminator}
\end{equation}
\begin{equation}
    \theta_{\textbf{T}^{s_k},f} \leftarrow \theta_{\textbf{T}^{s_k},f}
    - \eta_{T^{s_k}} \left( \frac{\partial \mathcal{L}_y^{s_k}}{\partial \theta_{\textbf{T}^{s_k},f}}
    - \lambda \frac{\partial \mathcal{L}_d^{s_k}}{\partial \theta_{\textbf{T}^{s_k},f}} \right),
\label{eq:update-encoder-1}
\end{equation}
where $\eta_{T^{s_k}}$ is the learning rate and $\lambda$ controls the strength of the adversarial signal. The negative sign in front of the discriminator gradient in Eq.~\eqref{eq:update-encoder-1} encourages the encoder to produce domain-invariant features that confuse the discriminator. We set $ \lambda = \frac{2}{1+\exp(-10\rho)} - 1$, where $\rho$ tracks the training progress, following standard practices in domain-adversarial learning~\cite{ganin2016domain,omidi2024unsupervised}. \\


\noindent{\bf Extension to classification and detection.}
While the formulation above is presented for segmentation, the same multi-dataset training strategy applies naturally to classification and object detection. The domain-adversarial mechanism remains unchanged: the encoder strives to produce domain-invariant representations, while the task-specific head (segmentation decoder, classifier, or detection head) is optimized using the appropriate supervised loss for each domain. For completeness, we summarize below the supervised loss functions for the additional tasks considered in this work. \\

\noindent \textit{Classification.} For classification domains, the supervised loss is the standard cross-entropy loss over class labels:
   \begin{equation}
        \mathcal{L}^{t}
        = \sum_{i=1}^{b} \text{CE}\!\left(\hat{p}^t_i, y^t_i\right),
    \end{equation}
    where $\hat{p}^t_i$ is the predicted class-probability vector and $y^t_i$ is the ground-truth label. \\

\noindent \textit{Object detection.} For detection domains using Faster R-CNN~\cite{ren2016fasterrcnnrealtimeobject}, the supervised loss combines a classification term and a Smooth~$L_1$ bounding-box regression term:
    \begin{equation}
        \mathcal{L}^{t}
        = \sum_{i=1}^{b} \Big(
            \mathcal{L}_{\text{cls}}\!(\hat{Y}^t_i, Y^t_i)
            +
            \mathcal{L}_{\text{reg}}\!(\hat{Y}^t_i, Y^t_i)
        \Big),
    \end{equation}
    where $\hat{Y}^t_i$ and $Y^t_i$ denote predicted and ground-truth bounding boxes and class labels.

For domains using RF-DETR ~\cite{robinson2025rfdetrneuralarchitecturesearch}, we employ a combination of classification, $L_1$ regression, and GIoU \cite{rezatofighi2019} loss (improves upon IoU by incorporating a penalty based on the smallest enclosing box, ensuring stable optimization even for non-overlapping predictions):
    \[
        \mathcal{L}^{t}
        = \sum_{i=1}^{b} \Big(
            \mathcal{L}_{\text{cls}}\!(\hat{Y}^t_i, Y^t_i)
            +
            \mathcal{L}_{\text{reg}}\!(\hat{Y}^t_i, Y^t_i)
            +
            \mathcal{L}_{\text{GIoU}}(\hat{Y}^t_i, Y^t_i)
        \Big),
    \]
    where $\hat{Y}^t_i$ and $Y^t_i$ again denote predicted and ground-truth annotations.
    
\medskip
In all cases, the supervised loss $\mathcal{L}^{t}$ replaces the segmentation term in Eq.~\eqref{eq:loss_divided_1}, whereas the discriminator loss $\mathcal{L}^{s_k}_d$ remains unchanged. This makes our teacher–student formulation task-agnostic and directly applicable to segmentation, classification, and object detection without altering the underlying domain-adversarial training scheme.

\vspace{2mm}
\subsection{Stage 2: Joint teacher via multi-level feature fusion}
\label{stage2-joint-teacher}

Once the target teacher $\mathcal{T}^t$ and the source teachers $\{\mathcal{T}^{s_k}\}_{k=1}^m$ are trained, we construct a \emph{joint teacher model} $\mathcal{T}_*$ by integrating their multi-level features (Stage 2 in Figure~\ref{fig:teacher-fusion}, Algorithm~\ref{alg_CDTS}, stage 2, lines~20--29). Each teacher architecture is composed of an encoder and a bottleneck that extract hierarchical representations at different spatial resolutions. The encoders capture fine-grained structures and high-level semantics, while the bottleneck encodes global context.

For each resolution level of the encoder, we collect the feature maps from the target teacher and all source teachers, and fuse them using cross-attention blocks that allow the model to aggregate complementary information from different domains. The fused representations are concatenated along the channel dimension and passed through convolutional residual blocks (with appropriate downsampling or upsampling) to harmonize their scale and statistics. The same procedure is applied at the bottleneck level to obtain a fused global representation.

In all experiments, the encoders and bottleneck layers of $\mathcal{T}^t$ and $\{\mathcal{T}^{s_k}\}$ are kept \emph{frozen} during fusion. A new task-specific head (segmentation decoder, classifier, or detection head) is then attached on top of the fused encoder–bottleneck backbone and trained from scratch using only the target dataset $\mathbf{D}^t$. This design preserves the domain-specific knowledge captured by each teacher while allowing the joint teacher $\mathcal{T}_*$ to specialize to the target task through its newly trained head.


\medskip
\noindent \textit{Classification.} For classification tasks, each teacher produces a high-level feature vector before its classification head. During joint fusion, the teacher encoders are kept frozen, and their embeddings are first passed through small adapter modules 
(e.g., $1 \times 1$ convolutions or linear projections) that map them into a shared representation space of dimension \texttt{jointch}. The adapted embeddings are concatenated and processed by a sequence of Fusion Attention blocks that refine cross-model interactions. The first block reduces the concatenated feature dimensionality to 
\texttt{jointch}, while subsequent blocks maintain this size. The output of the final Fusion Attention block is fed into a new  classification head trained from scratch on the target dataset.  Only the adapters, the Fusion Attention blocks, and the classifier are updated during this stage; all teacher encoders remain frozen. \\

\noindent \textit{Object detection.} For detection tasks, each teacher provides a set of multi-scale feature maps corresponding to its backbone. All teacher backbones are frozen, and fusion is performed independently at each resolution.  For an input image, we collect the feature maps from all teachers at  a given scale, concatenate them along the channel dimension, and  project them back to a fixed number of channels using a $1 \times 1$  convolution, followed by normalization and non-linearity.  This produces a harmonized set of fused multi-scale features that can  be consumed by any downstream detection head. We instantiate this generic fusion backbone in two detector families.  For Faster~R-CNN\cite{ren2016fasterrcnnrealtimeobject}, we replace the standard backbone with a  \emph{FusionBackbone} that wraps several frozen teacher backbones and  returns a dictionary of fused feature maps compatible with the region  proposal and detection heads. For RF-DETR \cite{robinson2025rfdetrneuralarchitecturesearch}, we construct an RF-DETR joint teacher in which the early  backbone stages are replaced by the fusion module, and the fused  features are subsequently processed by the transformer encoder–decoder and prediction heads. During training, only the fusion layers and the trainable components  of the detection head are updated, while all teacher backbones remain frozen.

\medskip
In all three cases (segmentation, classification, and detection) the fused backbone of $\mathcal{T}_*$ provides a unified representation that  aggregates complementary information from multiple teachers, enabling  the joint teacher to adapt effectively to the target domain while  remaining architecturally compatible with a wide range of task-specific 
heads.

\vspace{2mm}
\subsection{Stage 3: Multi-level knowledge distillation to the student}

The final stage distills knowledge from the joint teacher $\mathcal{T}_*$ into a compact student $\mathcal{S}^t$ (Algorithm~\ref{alg_CDTS}, stage 3, lines~30--41). The student is initialized with random weights and trained end-to-end, with all parameters updated. During training, we extract intermediate features from all encoder and bottleneck layers of $\mathcal{T}_*$ and use them to guide the student’s representations. The overall training objective is:
\begin{equation}
    \mathcal{L}_{\text{total}} = \alpha \,\mathcal{L}_{\text{task}} 
    + cf \,\big(\beta \,\mathcal{L}_{\text{Con}} 
                  + \gamma \,\mathcal{L}_{\text{FAlgn}} 
                  + \delta \,\mathcal{L}_{\text{CosSim}}\big),
\label{eq:student_loss}
\end{equation}
where $\mathcal{L}_{\text{task}}$ is the task-specific loss 
and $\mathcal{L}_{\text{Con}}$, $\mathcal{L}_{\text{FAlgn}}$, and $\mathcal{L}_{\text{CosSim}}$ are distillation terms:

\begin{itemize}
    \item $\mathcal{L}_{\text{task}}$. The segmentation task loss, $\mathcal{L}_{DiceBCE}$, combines Dice loss and Binary Cross-Entropy (BCE) loss to ensure accurate segmentation and pixel-wise classification:
    
    \begin{equation}
        \footnotesize
        \hspace{0pt}
        \begin{array}{l}
        \mathcal{L}_{\text{DiceBCE}} = \mathcal{L}_{\text{Dice}} + \mathcal{L}_{\text{BCE}}, \\[2mm]
        \mathcal{L}_{\text{Dice}} = 1 - \frac{2 \displaystyle\sum\limits_{i=1}^{N} \hat{y}_i y_i}{\displaystyle\sum\limits_{i=1}^{N} \hat{y}_i + \displaystyle\sum\limits_{i=1}^{N} y_i}, \text{ and } 
        \mathcal{L}_{\text{BCE}} = \frac{1}{N} \displaystyle\sum\limits_{i=1}^{N} \text{BCE}(\hat{y}_i, y_i),
        \end{array}
    \end{equation}
    
    \noindent where $y_i$ is the ground truth and $\hat{y}_i$ is the prediction.

    \item $\mathcal{L}_{Con}$. Contrastive loss~\cite{pmlr-v37-ganin15} aligns teacher and student probability distributions by minimizing the Kullback–Leibler (KL) divergence between their similarity distributions, encouraging robust feature learning:

    {\footnotesize	
    \setlength{\abovedisplayskip}{6pt}
    \setlength{\belowdisplayskip}{6pt}
    \begin{equation}
        \mathcal{L}_{Con} = \frac{1}{N} \sum_{i=1}^{N} \text{KL} \left( \text{softmax} \left( \frac{S_i}{t} \right) \parallel \text{softmax} \left( \frac{T_i}{t} \right) \right),
    \end{equation}
    }
    
    \noindent where $S_i$ and $T_i$ are similarity matrices for the student and teacher, respectively, and $t$ is a temperature scaling factor.  

    \item $\mathcal{L}_{FAlgn}$. Feature alignment loss~\cite{pmlr-v37-ganin15, DBLP:journals/corr/abs-2006-12770} minimizes discrepancies between teacher and student feature maps by reducing the mean squared error between them, promoting consistent representations:
    
    \begin{equation}
        \mathcal{L}_{FAlgn} = \frac{1}{N} \sum_{i=1}^{N} \| \hat{f}_i - f_i \|_2^2,
    \end{equation}

    \noindent where $\hat{f}_i$ and $f_i$ are feature maps from the student and teacher models, with the teacher's feature maps aligned using a $1 \times 1$ convolution.

    \item $\mathcal{L}_{CosSim}$. Cosine similarity loss~\cite{1467314} aligns teacher and student feature vectors directionally in high-dimensional space by maximizing their cosine similarity, encouraging directional consistency:

    \begin{equation}
        \mathcal{L}_{CosSim} = \frac{1}{N} \sum_{i=1}^{N} \left( 1 - \frac{\hat{f}_i \cdot f_i}{\|\hat{f}_i\|_2 \|f_i\|_2} \right),
    \end{equation}

    \noindent where $\hat{f}_i$ and $f_i$ are the student and teacher feature maps, respectively.
    
\end{itemize}

The scalar coefficients $\alpha$, $\beta$, $\gamma$, and $\delta$ control the relative importance of the different losses, and $cf$ is a \emph{curriculum factor} that increases over time. At the beginning of training, $cf$ is small and the optimization focuses primarily on the task loss $\mathcal{L}_{\text{task}}$, ensuring that the student first learns to solve the primary task. As training progresses, $cf$ gradually increases, giving more weight to the distillation terms and allowing the student to better align its internal representations with those of the joint teacher.

This curriculum-based distillation strategy encourages stable optimization: the student initially learns a reasonable solution driven by ground-truth supervision and then progressively refines its features to match the richer, multi-source representation space of $\mathcal{T}_*$. In practice, we observe that this leads to improved generalization across domains and consistent gains over both dataset-specific models and multi-head multi-dataset baselines, for segmentation, classification, and detection alike.
Table~\ref{tab:combined-1} presents the hyperparameter values ($\alpha, \beta, \gamma, \delta$, and $cf$) used in our training setup. \\

\noindent{\bf Extension to classification and detection.}
The distillation mechanism described above is task-agnostic and extends naturally beyond segmentation. The key idea remains unchanged: the student receives supervision from  both the task-specific ground-truth loss and the intermediate  representations produced by the joint teacher $\mathcal{T}_*$. Across tasks, only the definition of the supervised component $\mathcal{L}_{\text{task}}$ changes, while the distillation losses 
($\mathcal{L}_{\text{Con}}$, $\mathcal{L}_{\text{FAlgn}}$, $\mathcal{L}_{\text{CosSim}}$) and the curriculum mechanism remain identical. For classification, distillation operates on teacher logits and  multi-level embeddings; for detection, it operates on multi-scale feature maps and bounding-box predictions.\\

\noindent{\bf{Classification.}}
For image-level classification, we instantiate the generic objective in Eq.~\eqref{eq:student_loss} with a MedViT-based student classifier $\mathcal{S}^t$ that receives the same images as the joint teacher $\mathcal{T}_*$. In this setting, the task loss $\mathcal{L}_{\text{task}}$ is represented by $\mathcal{L}_{\text{CE}}$, a class-weighted cross-entropy between student predictions and ground-truth labels, with dynamic class weights to mitigate label imbalance:

\begin{equation}
    \begin{split}
        \mathcal{L}_{\text{CE}}
        &= - \sum_{c=1}^{C} w_c\, y_c \log p_c \,,
    \end{split}
\end{equation}
\noindent
where \(w_c = \tfrac{1}{f_c + \epsilon}\) are dynamic class weights derived from class frequencies \(f_c\), and \(p_c = \mathcal{S}^t(x)_c\) denotes the student’s predicted probability for class \(c\).

Of the three Stage~3 distillation terms: $\mathcal{L}_{\text{Con}}$, $\mathcal{L}_{\text{FAlgn}}$, and $\mathcal{L}_{\text{CosSim}}$, defined on multi-level segmentation features, we retain for the classification task only $\mathcal{L}_{\text{FAlgn}}$ and $\mathcal{L}_{\text{CosSim}}$. Concretely, we select a set of fused teacher feature maps and their student counterparts across $K$ layers, spatially pool them to obtain per-image embeddings, and aggregate these embeddings into the vectors $\hat{f}_i$ and $f_i$ used in Eq.~\eqref{eq:student_loss}. This yields (i) map-based feature alignment between corresponding intermediate representations, (ii) cosine-based alignment of pooled descriptors, and (iii) batch-wise contrastive alignment of their relational structure, all governed by the same curriculum factor $cf$.

In addition to these representation-level signals, we also introduce a \emph{logit-level} distillation term that transfers the teacher’s dark knowledge~\cite{hinton2015distilling}. Let $z_i^{(T)}$ and $z_i^{(S)}$ denote the teacher and student logits for sample $i$, and $t$ a temperature parameter. 
Following~\cite{hinton2015distilling}, we define:
\begin{equation}
    \mathcal{L}_{\text{KLD}}
    =
    \frac{1}{N}\sum_{i=1}^N 
    \mathrm{KL}\Big(
        \mathrm{softmax}\big(\tfrac{z_i^{(T)}}{t}\big)
        \,\Big\|\,
        \mathrm{softmax}\big(\tfrac{z_i^{(S)}}{t}\big)
    \Big)\, t^2,
\end{equation}
and include $\mathcal{L}_{\text{KLD}}$ inside the distillation part of Eq.~\eqref{eq:student_loss}, with its own scalar weight. Overall, the classification objective can be written as:
\begin{equation}
\begin{aligned}
    \mathcal{L}_{\text{total}}
    &=
    \alpha\,\mathcal{L}_{\text{task}}
    +
    cf \Big(\gamma \,\mathcal{L}_{\text{FAlgn}} 
      + \delta \,\mathcal{L}_{\text{CosSim}}
      + \kappa \,\mathcal{L}_{\text{KLD}}
    \Big),
\end{aligned}
\label{eq:cls_student_loss}
\end{equation}

\noindent where $\alpha$, $\gamma$, $\delta$, and $\kappa$ are scalar weights. Thus, the classification student benefits jointly from (i) hard-label supervision via $\mathcal{L}_{\text{task}}$, (ii) soft teacher predictions via $\mathcal{L}_{\text{KLD}}$, and (iii) multi-level feature and relational alignment via $\mathcal{L}_{\text{FAlgn}}$ and $\mathcal{L}_{\text{CosSim}}$.

\noindent \\
\noindent{\bf{Object detection.}}
For object detection, we again reuse the generic distillation objective in Eq.~\eqref{eq:student_loss}, but instantiate it with a student detector $\mathcal{S}^t$ (either Faster~R-CNN or RF-DETR) and detection-specific supervision. Here, the task loss $\mathcal{L}_{\text{task}}$ is the standard detection loss ($\mathcal{L}_{\text{cls}}$ classification and $\mathcal{L}_{\text{box}}$ box regression, including RPN/ROI terms for Faster~R-CNN or Hungarian-matched losses for RF-DETR), computed only from the student outputs:
\begin{itemize}
    \item Faster~R-CNN: 
    \begin{equation}
        \small
        \hspace{0pt}
        \begin{array}{l}
        \mathcal{L}_{\text{task}} = \mathcal{L}_{\text{cls}} + \mathcal{L}_{\text{box}} \\[2mm]
        \mathcal{L}_{\text{cls}} = \mathcal{L}_{\text{RPN-cls}} + \mathcal{L}_{\text{ROI-cls}} \quad
        \mathcal{L}_{\text{box}} = \mathcal{L}_{\text{RPN-reg}} + \mathcal{L}_{\text{ROI-reg}} \\[2mm]
        \mathcal{L}_{\text{RPN-cls}} = - \displaystyle\sum\limits_{i=1}^{N_{\text{rpn}}} y_i^{\text{rpn}} \log p_i^{\text{rpn}} \\[2mm]
        \mathcal{L}_{\text{RPN-reg}} = \displaystyle\sum\limits_{i=1}^{N_{\text{rpn}}} y_i^{\text{rpn}}\, \mathrm{SmoothL1}(b_i - \hat{b}_i) \\[2mm]
        \mathcal{L}_{\text{ROI-cls}} = - \displaystyle\sum\limits_{j=1}^{N_{\text{roi}}} y_j^{\text{roi}} \log p_j^{\text{roi}} \\[2mm]
        \mathcal{L}_{\text{ROI-reg}} = \displaystyle\sum\limits_{j=1}^{N_{\text{roi}}} y_j^{\text{roi}}\, \mathrm{SmoothL1}(b_j - \hat{b}_j),
        \end{array}
    \end{equation}

    \noindent
    where \(y_i^{\text{rpn}}\) and \(y_j^{\text{roi}}\) are the ground-truth labels for anchors and ROI proposals,  
    \(p_i^{\text{rpn}}\) and \(p_j^{\text{roi}}\) are the predicted classification probabilities,  
    and \(b_i, \hat{b}_i, b_j, \hat{b}_j\) are the ground-truth and predicted box coordinates at RPN and ROI levels.
    \item RF-DETR:
    \begin{equation}
        \small
        \hspace{0pt}
        \begin{array}{l}
        \mathcal{L}_{\text{task}} = \mathcal{L}_{\text{cls}} + \mathcal{L}_{\text{box}} \\[2mm]
        \mathcal{L}_{\text{cls}} = \displaystyle\sum_{(i,j)\in \pi} \bigl(- y_j \log p_i\bigr) \\[2mm]
        \mathcal{L}_{\text{box}} = \displaystyle\sum_{(i,j)\in \pi} 
        \Big(
        \lambda_1 \| b_i - \hat{b}_j \|_1
        + \lambda_2 \bigl(1 - \mathrm{IoU}(b_i, \hat{b}_j)\bigr)
        \Big),
        \end{array}
    \end{equation}

\noindent
where \(\pi\) is the optimal Hungarian assignment between predicted and ground-truth boxes,  
\(y_j\) are the ground-truth labels, \(p_i\) are the student predictions,  
\(b_i\) and \(\hat{b}_j\) are the predicted and ground-truth box coordinates, \(\lambda_1, \lambda_2\) are scalar weights for L1 and IoU regression terms.

\end{itemize}

The three distillation terms from Stage~3 are applied to multi-scale detection features. We treat the backbone feature maps of teacher and student at all levels as the intermediate representations used in $\mathcal{L}_{\text{FAlgn}}$, and derive pooled backbone descriptors to form the embeddings $\hat{f}_i$ and $f_i$ that enter $\mathcal{L}_{\text{CosSim}}$ and $\mathcal{L}_{\text{Con}}$. In this way, the contrastive, feature-alignment, and cosine-similarity losses encourage the student backbone to reproduce the teacher’s multi-scale representation geometry, while the curriculum factor $cf$ again controls the strength of these signals over training.

In addition to these representation-level signals, we  augment Eq.~\eqref{eq:student_loss} with additional distillation terms operating at the level of attention maps, head logits, head embeddings:
\begin{itemize}
    \item $\mathcal{L}_{\text{Att}}$. Attention transfer distillation ~\cite{Zagoruyko2017Attention}, computes spatial attention maps for both the teacher and the student by summing the squared activations across channels, for each backbone feature map, as follows:    
    \begin{equation}
        A_{i,\ell} = \sum_c f_{i,\ell}^{(c)2}, 
        \qquad 
        \hat{A}_{i,\ell} = \sum_c \hat{f}_{i,\ell}^{(c)2},
    \end{equation}
    normalize them,
    \begin{equation}
        \tilde{A}_{i,\ell}
        = \frac{A_{i,\ell}}{\|A_{i,\ell}\|_2},
        \qquad
        \tilde{\hat{A}}_{i,\ell}
        = \frac{\hat{A}_{i,\ell}}{\|\hat{A}_{i,\ell}\|_2},
    \end{equation}
    and penalize their discrepancy with an $L_2$ loss:
    \begin{equation}
        \mathcal{L}_{\text{Att}}
        =
        \frac{1}{N|\mathcal{L}|}
        \sum_{i=1}^N \sum_{\ell \in \mathcal{L}}
        \big\| \tilde{\hat{A}}_{i,\ell} - \tilde{A}_{i,\ell} \big\|_2^2,
    \end{equation}
    
    \noindent where $f_{i,\ell}^{(c)}$ and $\hat{f}_{i,\ell}^{(c)}$ are the teacher and student feature maps at channel $c$ for layer $\ell$ and image $i$, $A_{i,\ell}$ and $\hat{A}_{i,\ell}$ are their spatial energy maps, and $\tilde{A}_{i,\ell}$, $\tilde{\hat{A}}_{i,\ell}$ are the normalized versions. This encourages the student to attend to similar spatial regions as the teacher, independently of feature scale.

    \item $\mathcal{L}_{\text{ClsKD}}$. Head-level logit distillation~\cite{hinton2015distilling}, computes softened class probability distributions via a softmax function with temperature scaling over regions or queries in each image, followed by the computation of the Kullback-Leibler divergence across all regions/queries as:

    \begingroup
    \footnotesize
    \setlength{\abovedisplayskip}{6pt}
    \setlength{\belowdisplayskip}{6pt}
    \begin{equation}
        \mathcal{L}_{\text{ClsKD}}
        =
        \frac{1}{\sum_i M_i}
        \sum_{i=1}^N \sum_{j=1}^{M_i} 
        \mathrm{KL}\!\left(
            \mathrm{softmax}\!\left(\tfrac{z_{i,j}^{(T)}}{t}\right)
            \,\Big\|\,
            \mathrm{softmax}\!\left(\tfrac{z_{i,j}^{(S)}}{t}\right)
        \right),
    \end{equation}
    \endgroup    
    
    \noindent where $z_{i,j}^{(T)}$ and $z_{i,j}^{(S)}$ are the teacher and student logits for region/query $j$ in image $i$, $M_i$ is the number of proposals/queries, and $t$ is the temperature. This loss transfers the teacher’s class-level knowledge to the student at the proposal/query level.

    \item $\mathcal{L}_{\text{ROIKD}}$. Head-level feature alignment~\cite{Romero2015Fitnets}, aligns teacher and student representations by minimizing the squared $\ell_2$ distance between their respective embeddings, $h_{i,j}^{(T)}$ and $h_{i,j}^{(S)}$, obtained via ROI pooling (Faster~R-CNN) or decoder queries (RF-DETR):
    
    \begin{equation}
        \mathcal{L}_{\text{ROIKD}}
        =
        \frac{1}{\sum_i M_i} \sum_{i=1}^N \sum_{j=1}^{M_i}
        \big\| h_{i,j}^{(S)} - h_{i,j}^{(T)} \big\|_2^2,
    \end{equation}
    where $h_{i,j}^{(S)}$ and $h_{i,j}^{(T)}$ are the student and teacher head-level features for region/query $j$ in image $i$, and $M_i$ is the number of regions/queries in image $i$. This enforces consistency in the high-level semantic space directly used for object-level predictions and complements the backbone-level feature alignment from Stage~3.
\end{itemize}

All three detection-specific terms are added to the distillation part of Eq.~\eqref{eq:student_loss} via Eq.~\eqref{eq:det_student_loss}, each with its own scalar weight and modulated by the same curriculum factor $cf$. As a result, the detection student is guided not only by its supervised detection loss, but also by the teacher’s multi-scale backbone representations, spatial attention patterns, and head-level logits and embeddings, leading to improved robustness and accuracy across domains. Overall, the learning objective for detection is defined as:
\begin{equation}
\begin{aligned}
    \mathcal{L}_{\text{total}}
    &=
    \alpha\,\mathcal{L}_{\text{task}}
    +
    cf \Big(
        \beta \,\mathcal{L}_{\text{Con}}
      + \gamma \,\mathcal{L}_{\text{FAlgn}} 
      + \delta \,\mathcal{L}_{\text{CosSim}} \\
      &\qquad\qquad\quad
      + \eta \,\mathcal{L}_{\text{Att}}
      + \zeta \,\mathcal{L}_{\text{ClsKD}}
      + \rho \,\mathcal{L}_{\text{ROIKD}}
    \Big),
\end{aligned}
\label{eq:det_student_loss}
\end{equation}
where $\alpha, \beta, \gamma, \delta, \eta, \zeta$, and $\rho$ are scalar weights. Consequently, the detection student is trained jointly with (i) hard-label supervision via $\mathcal{L}_{\text{task}}$, and (ii) feature-level guidance via the Stage~3 losses: $\mathcal{L}_{\text{Con}}$, $\mathcal{L}_{\text{FAlgn}}$, $\mathcal{L}_{\text{CosSim}}$, $\mathcal{L}_{\text{Att}}$, $\mathcal{L}_{\text{ClsKD}}$, and $\mathcal{L}_{\text{ROIKD}}$.

In the following, we report our experimental evaluation across three tasks: segmentation, classification, and object detection. Because segmentation constitutes the original and primary application of our framework, we begin with a detailed analysis of segmentation performance.



\section{Segmentation Evaluation}

Segmentation represents the core application setting of our framework and serves as a challenging testbed for evaluating cross-domain generalization. Across six datasets drawn from different modalities and clinical conditions, we examine whether multi-dataset distillation enables the student model to surpass dataset-specific and multi-dataset baselines. We first introduce the datasets and baselines before presenting quantitative and qualitative results.

\subsection{Datasets}


\noindent\textbf{BrainMetShare}~\cite{grovik2020deep} is a collection of high-resolution pre- and post-contrast MRI sequences from patients with at least one confirmed brain metastasis.  The dataset includes 105 patients for training and 51 for testing, each with four 3D MRI sequences: T1 pre-contrast, T1 post-contrast, T1 gradient-echo post-contrast, and T2 FLAIR post-contrast. Expert-annotated binary segmentation masks of metastatic lesions are provided  for the training set. We use the post-contrast T2 FLAIR sequence in all experiments.


\medskip
\noindent\textbf{ISLES 2022}~\cite{hernandez2022isles,de2024robust} is a multi-center dataset 
containing 400 cases (250 training, 150 testing) of acute and sub-acute ischemic stroke. The scans exhibit strong variability in lesion appearance, size, and spatial distribution, with an average of 9{,}289 disconnected ischemic regions per case and a maximum of 126 regions. Data were acquired using three different imaging devices, making the dataset a challenging benchmark for cross-domain generalization. The available modalities include DWI, ADC, and FLAIR; in our experiments we use DWI, the clinical standard for ischemic core assessment.


\medskip
\noindent\textbf{BraTS 2020}~\cite{menze2014multimodal,bakas2017advancing,bakas2018identifying} comprises pre-operative MRI scans of patients diagnosed with low- or high-grade gliomas. The dataset contains 369 training cases, 125 validation cases, and 166 test cases. Each subject is scanned with four MRI sequences: T1-weighted, T1-weighted post-contrast, T2-weighted, and T2-weighted FLAIR. Ground truth annotations differentiate several tumor subregions, including contrast-enhancing tumor, necrotic core, and peritumoral edema. To maintain consistency with BrainMetShare, we use the post-contrast T2-weighted FLAIR modality.


\medskip
\noindent\textbf{Lung MSD}~\cite{antonelli2022medical} contains pre-operative thin-section CT scans from 96 patients with non-small cell lung cancer. The main challenge stems from the relatively small tumor regions embedded in a large thoracic field of view.


\medskip
\noindent\textbf{LiTS 2017}~\cite{bilic2023liver} includes CT volumes of primary and secondary liver tumors collected across seven hospitals and research centers. The dataset comprises 131 volumes for training and validation and an additional 70 test cases. Tumors vary widely in size, shape, and lesion-to-background contrast (hyper- or hypo-dense), making segmentation particularly difficult.


\medskip
\noindent\textbf{KiTS 2019}~\cite{heller2019kits19} provides multi-phase abdominal CT scans, segmentation masks, and accompanying clinical data for 300 patients who underwent nephrectomy for kidney tumors. Of these, 210 scans are used for training and validation and 90 are held out for testing. Annotations include detailed delineations of renal tumors and surrounding structures.

\medskip
\noindent\textbf{Dataset grouping.}  We categorize the six datasets by imaging modality: (1) \emph{MRI-based}: BrainMetShare, ISLES, and BraTS, covering brain metastases, ischemic stroke, and glioma segmentation; and (2) \emph{CT-based}: Lung MSD, LiTS, and KiTS, covering lung, liver, and kidney tumor segmentation tasks.This grouping highlights the cross-modality nature of our experimental setting and the heterogeneity addressed by our multi-dataset distillation framework.

\subsection{Teacher-student architectures}

Our framework is compatible with a broad class of encoder–decoder segmentation networks, enabling flexibility in choosing architectures that best suit each dataset. In our experiments, we consider two state-of-the-art models widely used in medical image segmentation, namely UNet~\cite{ronneberger2015u} and TResUNet~\cite{tomar2022transresunettransformerbasedresunet}. Both architectures are evaluated as teachers and as students within our distillation pipeline.

\medskip
\noindent\textbf{UNet}~\cite{ronneberger2015u}. UNet is a fully convolutional architecture designed specifically for biomedical image segmentation.  
It follows the characteristic U-shaped design, consisting of an encoder that progressively reduces spatial resolution to capture contextual and semantic information, and a decoder that restores spatial detail through successive upsampling operations.  
Skip connections are used to bridge corresponding encoder and decoder stages, allowing fine-grained spatial features to be combined with high-level contextual representations. This design enables accurate delineation of anatomical structures even when training data are limited.

\medskip
\noindent\textbf{TResUNet}~\cite{tomar2022transresunettransformerbasedresunet}.  
TResUNet extends the classical UNet architecture by integrating a ResNet50 backbone together with transformer-based self-attention and dilated convolutions. The ResNet50 encoder provides strong hierarchical feature extraction, while the transformer blocks enhance long-range dependency modeling—crucial for capturing global anatomical context.  
Dilated convolutions further enlarge the receptive field without loss of resolution. As in UNet, skip connections link the encoder and decoder stages, facilitating stable gradient flow and ensuring the preservation of spatial details.  Final segmentation masks are generated via a $1 \times 1$ convolution followed by a sigmoid activation for binary segmentation.

\medskip
These two architectures provide complementary strengths: UNet offers a lightweight and efficient baseline, while TResUNet incorporates transformer-based global reasoning.  
Evaluating both architectures within our teacher–student framework allows us to assess the generality and robustness of the proposed multi-dataset distillation approach across different model capacities and representational biases.

\subsection{Baselines}
\label{sec:segmentation_baselines}
We evaluate our approach against two categories of baselines designed to isolate the contribution of multi-dataset and cross-domain distillation.

\medskip
\noindent\textbf{Dataset-specific models.} These models use the same architecture as our student networks but are trained \emph{independently} on each dataset, from scratch and without any form of cross-dataset supervision.  They represent a standard fully supervised setting and provide a reference point for quantifying the benefit of incorporating knowledge from multiple domains.

\medskip
\noindent\textbf{Multi-head multi-dataset models.}
In this baseline, a shared backbone is trained jointly on all datasets, while a separate prediction head is assigned to each dataset.  During training, mini-batches from the different datasets are sampled in turn, and the corresponding losses are accumulated so that each dataset contributes equally before the shared backbone is updated. This setup evaluates the benefit of collaborative learning across datasets using a unified feature extractor, but without the explicit distillation or feature fusion mechanisms introduced in our method.

\medskip
\noindent Performance improvements over these two baselines demonstrate the effectiveness of our proposed approach in exploiting cross-domain information and integrating multi-dataset knowledge for improved segmentation accuracy.

\subsection{Experimental setup} 


We evaluate our framework on six datasets covering brain tumors (MRI/DWI) and organ tumors (CT). The experiments are designed to measure how effectively the proposed multi-dataset distillation improves segmentation performance across heterogeneous imaging domains.


\medskip
\noindent{\bf Dataset Splits and Evaluation Protocol.} For each dataset, we use only the portion with available ground-truth segmentation masks. This corresponds to \textit{MRI/DWI scans} from 724 patients (BrainMetShare, ISLES, BraTS) and \textit{CT scans} from 437 patients (Lung MSD, LiTS, KiTS).  
Each dataset is partitioned into $80\%$ training, $10\%$ validation, and $10\%$ testing. This uniform split allows a fair comparison between: (i) \textit{dataset-specific baselines}, trained independently on each dataset; (ii) \textit{student models}, trained via knowledge distillation from the joint teacher.

For each imaging modality (MRI/DWI or CT), we evaluate our framework in a \textit{target–source} configuration:  each dataset is treated once as the \emph{target}, while the remaining two datasets in the same modality serve as \emph{sources}.  This setup highlights the cross-domain transfer capabilities of our method.

\medskip
\noindent \textbf{Hyperparameter details.} 
All teacher and student networks are trained with a learning rate of $10^{-4}$ and a batch size of $8$ (see Table~\ref{tab:combined-1} for a complete overview of the hyperparameters). A learning rate reduction factor of $0.1$ is applied whenever the validation loss does not improve for five consecutive epochs. The segmentation loss weight is set to $\alpha = 0.5$, balancing direct supervision with the distillation losses. Temperature annealing is used for contrastive distillation, starting from $t_1 = 2$ and decaying according to $t_2 = \max\left(0.5,\; t_1 \times 0.9^{\text{epoch}/5}\right)$. The contrastive loss weight increases gradually from $\beta_1 = 0.1$ to $\beta_{\max} = 0.5$ as training progresses. Feature-alignment and cosine-similarity losses use weights $\gamma = 0.3$ and $\delta = 0.1$, respectively. We employ a curriculum factor $cf$ to progressively activate knowledge distillation: $cf = \min\left(1,\; \max\left(0,\; \frac{\text{epoch} - \text{warmup\_epochs}}{\text{ramp\_epochs}} \right)\right)$ with $\text{warmup\_epochs} = 5$ and $\text{ramp\_epochs} = 10$. This schedule ensures that the student first learns the supervised task before aligning with the joint teacher’s feature space. Models are trained for up to 300 epochs with early stopping based on validation performance.

\begin{table*}[t!]
\scriptsize 
\caption{Hyperparameters.}
\begin{tabular*}{\hsize}{@{\extracolsep{\fill}}lll@{}}
\toprule
\textbf{Hyperparameter} & \textbf{Description} & \textbf{Value} \\
\toprule
Learning rate ($\eta_{{\textbf{T}^{t}}}, \eta_{{\textbf{T}^{s_k}}}$) & Regulates parameter updates & \( 1e-4 \) \\ 
Batch size (b) & Number of training samples processed in one pass & \( 8 \) \\
Segmentation loss weight ($\alpha$) & Balances primary task and KD losses & \( 0.5 \) \\
Temperature scaling ($t$) & Sharpens distributions for contrastive loss & Start: \( t_1 = 2 \), Decay: \( t_2 = \max(0.5, t_1 \times 0.9^{\text{epoch}/5}) \) \\
Contrastive weight ($\beta$) & Gradual introduction of contrastive loss & Start: \( \beta_1 = 0.1 \), Max: \( \beta_{\max} = 0.5 \) \\
Feature map loss weight (\( \gamma \)) & Aligns teacher and student features & \( 0.3 \) \\
Similarity loss weight ($\delta$) & Contribution of the similarity loss & 0.1 \\ 
Curriculum factor & Dynamic curriculum for scheduling KD losses & \(cf = \)
\(\min\left(1, \max\left(0, \frac{epoch - \text{warmup\_epochs}}{\text{ramp\_epochs}}\right)\right)\) \\ 
Warmup epochs, ramp epochs & Control KD transition, disabling it, then increasing it & $warmup\_epochs=5$, $ramp\_epochs=10$ \\ 
\bottomrule
\end{tabular*}
\label{tab:combined-1}
\end{table*}

\medskip
\noindent {\bf Implementation details.} 
All segmentation, classification, and detection experiments are conducted on a workstation equipped with an NVIDIA GeForce RTX~3090 GPU, an Intel i9-10940X 3.3\,GHz CPU, and 128\,GB of RAM.

\subsection{Results} 
We assess segmentation performance using five standard metrics: Dice Similarity Coefficient (DSC), Precision, Recall, Intersection over Union (IoU), and the 95th percentile Hausdorff Distance (HD95). DSC quantifies the volumetric overlap between predicted and ground-truth masks, while Precision and Recall measure segmentation specificity and sensitivity, respectively.  IoU evaluates the ratio between the intersection and union of the predicted and ground-truth regions, and HD95 captures boundary accuracy by reporting the 95th percentile of the bidirectional surface distance. Table~\ref{tab:avg_performance} presents the average results for all baseline and student models across the MRI-based and CT-based datasets.  
We report performance for multiple teacher–student configurations using both UNet and TResUNet architectures, demonstrating the consistent improvements obtained through our multi-dataset distillation framework.


\begin{table*}[!htbp]
\centering
\scriptsize 
\setlength{\tabcolsep}{4pt} 
\caption{Average Performance of Baseline vs. Student Models for MRI and CT-based datasets using TResUNet and UNet.}
\begin{tabular*}{\hsize}{@{\extracolsep{\fill}}@{}c@{} c@{} c@{} c@{} c@{} c@{} c@{} c@{}}
\toprule
\textbf{Dataset Group} & \textbf{Model} & \textbf{Architecture} & \textbf{IoU ↑} & \textbf{DSC ↑} & \textbf{Recall ↑} & \textbf{Precision ↑} & \textbf{HD95 ↓} \\
\midrule
\multirow{2}{*}{MRI-based} 
& Multi-head  & TResUNet & 76.92 & 80.47 & 82.80 & 86.39 & 13.51 \\
& Dataset-specific   & TResUNet & 83.97 & 86.89 & 89.36 & 93.38 & 10.44 \\
& Joint teacher & TResUNet & 84.02 & 86.99 & 89.18 & 93.59 & 10.21 \\
& Student& TResUNet & \textbf{85.32} & \textbf{88.05} & \textbf{89.88} & \textbf{94.26} & \textbf{9.24} \\
\midrule
\multirow{2}{*}{CT-based} 
& Multi-head  & TResUNet & 79.91 & 82.55 & 83.48 & 89.17 & 12.55 \\
& Dataset-specific & TResUNet & 86.83 & 89.39 & 90.19 & \textbf{95.85} & 9.84 \\
& Joint teacher & TResUNet & 87.09 & 89.63 & 90.43 & 95.77 & 9.36 \\
& Student & TResUNet & \textbf{89.35} & \textbf{91.93} & \textbf{92.87} & 95.49 & \textbf{6.12} \\
\midrule
\midrule
\multirow{2}{*}{MRI-based } 
& Multi-head & UNet & 77.74 & 80.64 & 82.43 & 86.72 & 12.73 \\
& Dataset-specific & UNet & 84.21 & 87.07 & 88.99 & 93.90 & 10.03 \\
& Joint teacher & UNet & 84.11 & 86.97 & 88.69 & 93.69 & 10.15 \\
& Student & UNet & \textbf{85.63} & \textbf{88.49} & \textbf{89.69} & \textbf{94.53} & \textbf{8.73} 
\\
\midrule
\multirow{2}{*}{CT-based} 
& Multi-head & UNet & 77.49 & 80.51 & 82.05 & 87.18 & 14.38 \\
& Dataset-specific & UNet & 84.44 & 87.19 & 89.06 & 94.44 & 11.86 \\
& Joint teacher & UNet & 85.61 & 88.32 & 90.13 & 94.74 & 9.63 \\
& Student & UNet &  \textbf{89.22} & \textbf{91.32} & \textbf{93.22} & \textbf{95.41} & \textbf{6.55} \\
\midrule
\midrule
\multirow{2}{*}{MRI-based} 
& Dataset-specific & UNet & 84.21 & 87.07 & 88.99 & \textbf{93.90} & 10.03 \\
& Joint teacher & TResUNet & 84.02 & 86.99 & 89.18 & 93.59 & 10.21 \\
& Student  & UNet & \textbf{84.89} & \textbf{87.70} & \textbf{89.67} & 93.71 & \textbf{9.47} \\
\midrule
\multirow{2}{*}{CT-based} 
& Dataset-specific & UNet & 84.44 & 87.19 & 89.06 & 94.44 & 11.86 \\
& Joint teacher & TResUNet & 87.09 & 89.63 & 90.43 & \textbf{95.77} & 9.36 \\
& Student  & UNet & \textbf{88.30} & \textbf{90.70} & \textbf{91.66} & 95.59 & \textbf{6.95} \\
\bottomrule
\end{tabular*}
\label{tab:avg_performance}
\end{table*}

\medskip
\noindent\textbf{Students vs. baselines.}  
Across all datasets and architectures, the student models consistently outperform both the multi-head and dataset-specific baselines. 
For MRI-based tasks using TResUNet, the student achieves an average IoU improvement of approximately 1.3\%, accompanied by similar gains in DSC and recall, and a reduction of roughly 1.2 pixels in HD95.  
The improvements are even more substantial on CT-based datasets, where the student yields an average IoU increase of about 2.5\% along with consistent enhancements in the remaining metrics. 
A comparable trend is observed with the UNet architecture.  
On MRI datasets, the student improves IoU by around 1.4\% over the dataset-specific baseline, while on CT datasets the average gain rises to approximately 4.8\%.  
These improvements manifest not only in IoU and DSC but also in higher precision and lower HD95 values, reflecting more accurate and stable boundary predictions. Overall, the consistent performance gains across modalities and architectures highlight the strength of the proposed teacher–student framework in exploiting cross-domain information. The effect is particularly pronounced in CT-based segmentation tasks, where heterogeneity and lower soft-tissue contrast generally make cross-dataset transfer more challenging.

\medskip
\noindent \textbf{Performance of joint teachers.}
For completeness, Table~\ref{tab:avg_performance} also reports the performance of the joint teacher models.  
By fusing features from several source teachers, the joint teacher consistently surpasses the multi-head baseline across nearly all metrics and datasets and generally performs slightly better than the dataset-specific models.  
This confirms that aggregating domain-invariant representations from multiple sources yields a stronger and more versatile feature extractor.
However, the student models obtained through knowledge distillation invariably achieve the best results.  
Although the joint teacher benefits from cross-dataset feature fusion and domain adaptation, the distillation stage further tailors these representations to the target domain, producing a more compact and specialized model.  
This refinement is reflected in consistent improvements in IoU, DSC, and HD95 across both MRI-based and CT-based datasets, demonstrating the effectiveness of the distillation process in enhancing segmentation accuracy.

\medskip
\noindent\textbf{TResUNet vs. UNet.}
We next analyze the effect of architectural choice across both MRI-based and CT-based datasets. Overall, UNet and TResUNet exhibit comparable performance in terms of overlap-based metrics, indicating that the proposed framework is largely architecture-agnostic.

On MRI-based datasets, UNet consistently achieves slightly higher IoU and DSC scores than TResUNet, along with modest gains in Precision and Recall. Importantly, UNet also attains lower HD95 values in this setting, suggesting more accurate boundary localization for brain lesion segmentation. These results indicate that, for MRI data, the simpler convolutional design of UNet remains highly effective when combined with multi-dataset distillation.

In contrast, on CT-based datasets, the performance gap between the two architectures becomes less pronounced, and both models benefit substantially from the proposed teacher-student framework. While overlap metrics remain comparable, student models based on both UNet and
TResUNet achieve large reductions in HD95, reflecting improved boundary accuracy in challenging segmentation tasks.

To further assess architectural robustness, we additionally evaluate a cross-architecture configuration in which TResUNet is used for all teacher models and UNet serves as the student architecture. As reported in the final rows of Table~\ref{tab:avg_performance}, the resulting student performance remains consistent with the homogeneous-architecture settings, confirming that the proposed framework effectively transfers multi-dataset knowledge even when teacher and student architectures differ.
%

\medskip
\noindent\textbf{Per-dataset error analysis.} Tables~\ref{tab:all-tresunet-new}, \ref{tab:all-unet-new}, and \ref{tab:tresunet-unet-new} provide a detailed per-dataset breakdown of the segmentation results, using TResUNet-only, UNet-only, and mixed (TResUNet teacher – UNet student) architectures, respectively. This analysis reveals consistent trends across datasets, as well as
dataset-specific behaviors that are obscured by aggregate metrics.

On MRI data, the proposed distillation framework yields its strongest gains on \textit{BrainMetShare}, which is characterized by numerous small and sparsely distributed metastases. For both UNet and TResUNet architectures, the student improves IoU by more than 3 points compared to the dataset-specific baseline, while HD95 is reduced by
approximately 3 pixels. This indicates that multi-dataset fusion helps recover small lesions that are
frequently missed by models trained in isolation. On \textit{ISLES}, the improvements are more moderate but consistent across all metrics, reflecting the benefit of cross-dataset feature sharing in the presence of heterogeneous infarct patterns. In contrast, \textit{BraTS} exhibits a different behavior. For both UNet and TResUNet, the joint teacher underperforms the dataset-specific baseline, indicating that direct multi-dataset feature fusion is less effective for highly heterogeneous glioma segmentation. The student models consistently improve upon the joint teacher, recovering part of the performance loss, but remain slightly below the dataset-specific models by approximately 0.7--0.9 IoU/DSC points and with a modest increase in HD95. This pattern suggests that while knowledge distillation alleviates the over-regularization introduced at the joint teacher level, the infiltrative and multi-scale nature of gliomas limits the direct transfer of representations learned from metastasis and stroke datasets.

On CT data, the impact of distillation is more pronounced and is dominated by improvements in boundary accuracy. For \textit{Lung MSD} and \textit{LiTS}, the student achieves IoU gains of approximately +2.0 and +3.6 points, respectively, accompanied by sharp reductions in HD95 (between 3.5 and 4.8 pixels). These results indicate substantially crisper delineation of irregular lung
nodules and highly textured liver tumors. For \textit{KiTS}, the overlap improvements are more modest (around +1.9 IoU points), but the reduction in HD95 remains substantial (nearly 3 pixels), highlighting improved contour localization even when volumetric gains are limited.

The mixed TResUNet-teacher/UNet-student configuration
(Table~\ref{tab:tresunet-unet-new}) closely mirrors the trends observed in the homogeneous-architecture settings. In particular, CT-based datasets retain strong gains in both overlap and boundary metrics, confirming that the benefits of the proposed framework do not rely on matching teacher/student architectures.

\begin{table*}[!htbp]
\scriptsize
\setlength{\tabcolsep}{3pt}
\caption{Performance on MRI (BrainMetShare, ISLES, BraTS) and CT (Lung MSD, LiTS, KiTS) datasets using TResUNet as the architecture for all student, joint-teacher, and baseline models.}
\centering
\begin{tabular*}{\hsize}{@{\extracolsep{\fill}}@{}c@{} c@{} c@{} c@{} c@{} c@{} c@{} c@{} c@{} c@{} c@{} c@{} c@{} c@{} c@{} c@{}}
\toprule
\multirow{2}{*}{\textbf{Model}} & \multicolumn{5}{c}{\textbf{BrainMetShare}} & \multicolumn{5}{c}{\textbf{ISLES}} & \multicolumn{5}{c}{\textbf{BraTS}} \\
\cmidrule(lr){2-6} \cmidrule(lr){7-11} \cmidrule(lr){12-16}
& \textbf{IoU} & \textbf{DSC} & \textbf{Recall} & \textbf{Prec} & \textbf{HD95}
& \textbf{IoU} & \textbf{DSC} & \textbf{Recall} & \textbf{Prec} & \textbf{HD95}
& \textbf{IoU} & \textbf{DSC} & \textbf{Recall} & \textbf{Prec} & \textbf{HD95} \\
\midrule
$\text{Multi-head}$    & 75.33 & 78.15 & 81.74 & 85.31 & 13.24 & 74.21 & 79.11 & 79.48 & 87.16 & 15.07 & 81.23 & 84.14 & 87.19 & 86.71 & 12.22 \\
$\text{Dataset-specific}$      & 82.84 & 85.19 & 88.21 & 92.62 & 10.99 & 81.77 & 85.29 & 86.31 & 94.45 & 12.13 & \textbf{87.29} & \textbf{90.19} & \textbf{93.55} & \textbf{93.07} & \textbf{8.20} \\
$\text{Joint teacher}$ & 83.52 & 85.89 & 88.16 & 93.22 & 10.16 & 82.50 & 86.01 & 86.50 & 95.06 & 11.26 & 86.04 & 89.07 & 92.88 & 92.48 & 9.20 \\
$\text{Student}$       & \textbf{86.18} & \textbf{88.55} & \textbf{89.76} & \textbf{94.73} & \textbf{7.95} & \textbf{83.23} & \textbf{86.31} & \textbf{86.80} & \textbf{95.36} & \textbf{10.92} & 86.54 & 89.28 & 93.09 & 92.69 & 8.86 \\
\midrule
\midrule
\multirow{2}{*}{\textbf{Model}} & \multicolumn{5}{c}{\textbf{Lung MSD}} & \multicolumn{5}{c}{\textbf{LiTS}} & \multicolumn{5}{c}{\textbf{KiTS}} \\
\cmidrule(lr){2-6} \cmidrule(lr){7-11} \cmidrule(lr){12-16}
& \textbf{IoU} & \textbf{DSC} & \textbf{Recall} & \textbf{Prec} & \textbf{HD95}
& \textbf{IoU} & \textbf{DSC} & \textbf{Recall} & \textbf{Prec} & \textbf{HD95}
& \textbf{IoU} & \textbf{DSC} & \textbf{Recall} & \textbf{Prec} & \textbf{HD95} \\
\midrule
$\text{Multi-head}$    & 81.22 & 83.16 & 83.99 & 89.21 & 10.21  & 77.28 & 81.29 & 82.83 & 88.11 & 14.13 & 81.22 & 83.19 & 83.61 & 90.19 & 13.31 \\
$\text{Dataset-specific}$      & 88.06 & 90.53 & 90.86 & \textbf{96.48} & 7.71  & 84.71 & 88.13 & 89.02 & 94.49 & 11.01  & 87.72 & 89.50 & 90.69 & 96.58 & 10.81 \\
$\text{Joint teacher}$ & 88.56 & 90.83 & 91.16 & 95.93 & 6.78  & 84.80 & 88.30 & 89.25 & \textbf{94.65} & 10.78  & 87.92  & 89.75  & 90.88  & \textbf{96.73}  & 10.52 \\
$\text{Student}$       & \textbf{90.09} & \textbf{92.72} & \textbf{93.61} & 95.40 & \textbf{4.23}  & \textbf{88.35} & \textbf{91.42} & \textbf{92.46} & 94.63 & \textbf{6.17}  & \textbf{89.61}  & \textbf{91.65}  & \textbf{92.54}  & 96.44  & \textbf{7.95} \\
\bottomrule
\end{tabular*}
\label{tab:all-tresunet-new}
\end{table*}

\begin{table*}[!htbp]
\scriptsize 
\setlength{\tabcolsep}{3pt} 
\caption{Performance on MRI (BrainMetShare, ISLES, BraTS) and CT (Lung MSD, LiTS, KiTS) datasets using UNet as the architecture for all student, joint-teacher, and baseline models.}
\centering
\begin{tabular*}{\hsize}{@{\extracolsep{\fill}}@{}c@{} c@{} c@{} c@{} c@{} c@{} c@{} c@{} c@{} c@{} c@{} c@{} c@{} c@{} c@{} c@{}}
\toprule
\multirow{2}{*}{\textbf{Model}} & \multicolumn{5}{c}{\textbf{BrainMetShare}} & \multicolumn{5}{c}{\textbf{ISLES}} & \multicolumn{5}{c}{\textbf{BraTS}} \\
\cmidrule(lr){2-6} \cmidrule(lr){7-11} \cmidrule(lr){12-16}
& \textbf{IoU} & \textbf{DSC} & \textbf{Recall} & \textbf{Prec} & \textbf{HD95} 
& \textbf{IoU} & \textbf{DSC} & \textbf{Recall} & \textbf{Prec} & \textbf{HD95} 
& \textbf{IoU} & \textbf{DSC} & \textbf{Recall} & \textbf{Prec} & \textbf{HD95} \\
\midrule
$\text{Multi-head}$     & 75.11 & 77.28 & 80.28 & 85.47 & 14.01  & 75.77 & 78.87 & 80.21 & 86.21 & 15.02  & 82.34 & 85.77 & 86.79 & 88.48 & 9.15 \\
$\text{Dataset-specific}$       & 82.39 & 84.61 & 87.17 & 92.95 & 11.11  & 82.04 & 85.59 & 87.00 & 93.94 & 12.06  & \textbf{88.19} & \textbf{91.02} & \textbf{92.79} & \textbf{94.82} & \textbf{6.91} \\
$\text{Joint teacher}$  & 82.47 & 84.73 & 87.72 & 92.56 & 11.02  & 82.85 & 86.27 & 86.89 & 94.99 & 11.45  & 87.01 & 89.90 & 91.45 & 93.53 & 7.98 \\
$\text{Student}$        & \textbf{86.26} & \textbf{88.60} & \textbf{89.82} & \textbf{94.69} & \textbf{7.83}  & \textbf{83.05} & \textbf{86.47} & \textbf{87.27} & \textbf{95.02} & \textbf{11.13}  & 87.58 & 90.41 & 91.99 & 93.87 & 7.24 \\
\midrule
\midrule
\multirow{2}{*}{\textbf{Model}} & \multicolumn{5}{c}{\textbf{Lung MSD}} & \multicolumn{5}{c}{\textbf{LiTS}} & \multicolumn{5}{c}{\textbf{KiTS}} \\
\cmidrule(lr){2-6} \cmidrule(lr){7-11} \cmidrule(lr){12-16}
& \textbf{IoU} & \textbf{DSC} & \textbf{Recall} & \textbf{Prec} & \textbf{HD95} 
& \textbf{IoU} & \textbf{DSC} & \textbf{Recall} & \textbf{Prec} & \textbf{HD95} 
& \textbf{IoU} & \textbf{DSC} & \textbf{Recall} & \textbf{Prec} & \textbf{HD95} \\
\midrule
$\text{Multi-head}$    & 78.87 & 81.43 & 82.73 & 87.28 & 12.71 & 73.88 & 77.94 & 79.28 & 85.49 & 17.15  & 79.72 & 82.15 & 84.14 & 88.76 & 13.27 \\
$\text{Dataset-specific}$      & 85.45 & 88.13 & 89.97 & 94.71 & 9.87  & 80.96 & 84.64 & 86.88 & 92.69 & 14.77  & 86.92 & 88.79 & 90.33 & 95.93 & 10.93 \\
$\text{Joint teacher}$  & 85.93 & 88.51 & 90.34 & 94.98 & 9.01  & 83.87 & 87.55 & 89.63 & 93.22 & 9.36  & 87.04 & 88.91 & 90.41 & 96.01 & 10.53 \\
$\text{Student}$    & \textbf{89.79} & \textbf{92.36} & \textbf{93.67} & \textbf{95.14} & \textbf{5.02}  & \textbf{88.46} & \textbf{91.36} & \textbf{93.29} & \textbf{94.04} & \textbf{6.80}  & \textbf{89.42} & \textbf{90.23} & \textbf{92.71} & \textbf{97.05} & \textbf{7.83} \\
\bottomrule
\end{tabular*}
\label{tab:all-unet-new}
\end{table*}

\begin{table*}[!htbp]
\scriptsize 
\setlength{\tabcolsep}{3pt} 
\caption{Performance on MRI (BrainMetShare, ISLES, BraTS) and CT (Lung MSD, LiTS, KiTS) datasets using TResUNet as the teacher architecture and UNet as the student and baseline architecture.}
\centering
\begin{tabular*}{\hsize}{@{\extracolsep{\fill}}@{}c@{} c@{} c@{} c@{} c@{} c@{} c@{} c@{} c@{} c@{} c@{} c@{} c@{} c@{} c@{} c@{}}
\toprule
\multirow{2}{*}{\textbf{Model}} & \multicolumn{5}{c}{\textbf{BrainMetShare}} & \multicolumn{5}{c}{\textbf{ISLES}} & \multicolumn{5}{c}{\textbf{BraTS}} \\
\cmidrule(lr){2-6} \cmidrule(lr){7-11} \cmidrule(lr){12-16}
& \textbf{IoU} & \textbf{DSC} & \textbf{Recall} & \textbf{Prec} & \textbf{HD95} 
& \textbf{IoU} & \textbf{DSC} & \textbf{Recall} & \textbf{Prec} & \textbf{HD95} 
& \textbf{IoU} & \textbf{DSC} & \textbf{Recall} & \textbf{Prec} & \textbf{HD95} \\
\midrule
$\text{Multi-head}$     & 75.11 & 77.28 & 80.28 & 85.47 & 14.01  & 75.77 & 78.87 & 80.21 & 86.21 & 15.02  & 82.34 & 85.77 & 86.79 & 88.48 & 9.15 \\
$\text{Dataset-specific}$      & 82.39 & 84.61 & 87.17 & 92.95 & 11.11  & 82.04 & 85.59 & \textbf{87.00} & 93.94 & 12.06  & \textbf{88.19} & \textbf{91.02} & 92.79 & \textbf{94.82} & \textbf{6.91} \\
$\text{Joint teacher}$ & 83.52 & 85.89 & 88.16 & 93.22 & 10.16 & 82.50 & 86.01 & 86.50 & 95.06 & 11.26 & 86.04 & 89.07 & \textbf{92.88} & 92.48 & 9.20 \\
$\text{Student}$    & \textbf{85.67} & \textbf{88.04} & \textbf{89.83} & \textbf{93.89} & \textbf{8.36}  & \textbf{82.98} & \textbf{86.25} & 86.75 & \textbf{95.12} & \textbf{11.03}  & 86.02 & 88.81 & 92.42 & 92.12 & 9.01 \\
\midrule
\midrule
\multirow{2}{*}{\textbf{Model}} & \multicolumn{5}{c}{\textbf{Lung MSD}} & \multicolumn{5}{c}{\textbf{LiTS}} & \multicolumn{5}{c}{\textbf{KiTS}} \\
\cmidrule(lr){2-6} \cmidrule(lr){7-11} \cmidrule(lr){12-16}
& \textbf{IoU} & \textbf{DSC} & \textbf{Recall} & \textbf{Prec} & \textbf{HD95} 
& \textbf{IoU} & \textbf{DSC} & \textbf{Recall} & \textbf{Prec} & \textbf{HD95} 
& \textbf{IoU} & \textbf{DSC} & \textbf{Recall} & \textbf{Prec} & \textbf{HD95} \\
\midrule
$\text{Multi-head}$    & 78.87 & 81.43 & 82.73 & 87.28 & 12.71 & 73.88 & 77.94 & 79.28 & 85.49 & 17.15  & 79.72 & 82.15 & 84.14 & 88.76 & 13.27 \\
$\text{Dataset-specific}$      & 85.45 & 88.13 & 89.97 & 94.71 & 9.87  & 80.96 & 84.64 & 86.88 & 92.69 & 14.77  & 86.92 & 88.79 & 90.33 & 95.93 & 10.93 \\
$\text{Joint teacher}$ & 88.56 & 90.83 & 91.16 & \textbf{95.93} & 6.78  & 84.80 & 88.30 & 89.25 & \textbf{94.65} & 10.78  & 87.92  & 89.75  & 90.88  & \textbf{96.73}  & 10.52 \\
$\text{Student}$    & \textbf{89.02} & \textbf{91.43} & \textbf{92.53} & 95.62 & \textbf{5.12}  & \textbf{87.36} & \textbf{90.33} & \textbf{91.21} & 94.62 & \textbf{7.16}  & \textbf{88.51} & \textbf{90.34} & \textbf{91.23} & 96.54 & \textbf{8.56} \\
\bottomrule
\end{tabular*}
\label{tab:tresunet-unet-new}
\end{table*}

\medskip 
\noindent{\bf Wall clock-times.} Constructing the joint teacher (Stage 2) is the only costly step, taking roughly three times the per-iteration training time of a single-dataset baseline ($\approx$  0.012s vs. 0.004s). Once distilled, the student runs at baseline speed and memory, $\approx$ 0.0006s per image at inference, while delivering the accuracy gains reported in Table \ref{tab:avg_performance}.

\begin{figure*}[ht!]
    \centering
    \begin{minipage}{0.32\linewidth}
        \centering
        \includegraphics[width=\linewidth]{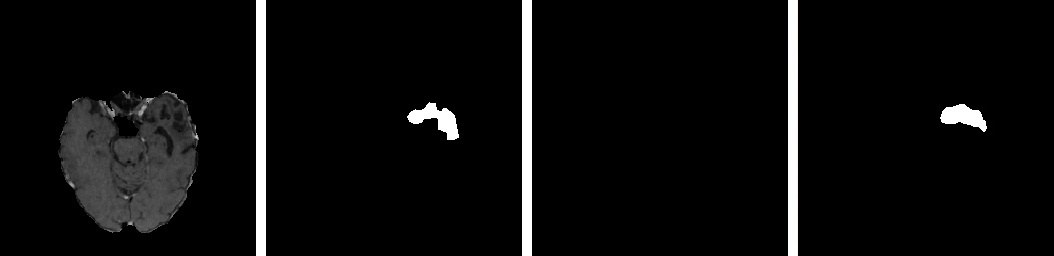}
        \vskip 0.08cm
        \includegraphics[width=\linewidth]{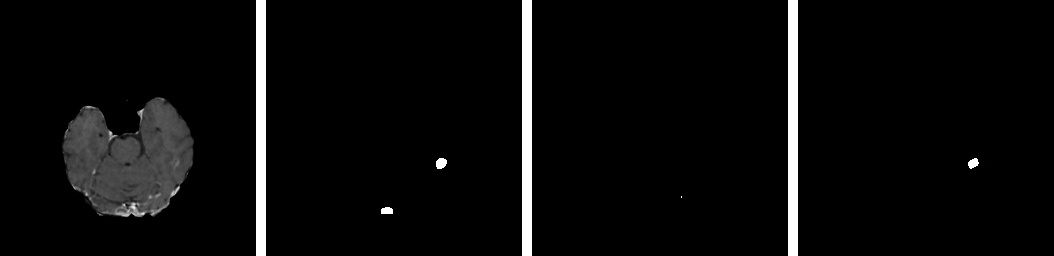}
    \end{minipage}
    \hfill
    \begin{minipage}{0.32\linewidth}
        \centering
        \includegraphics[width=\linewidth]{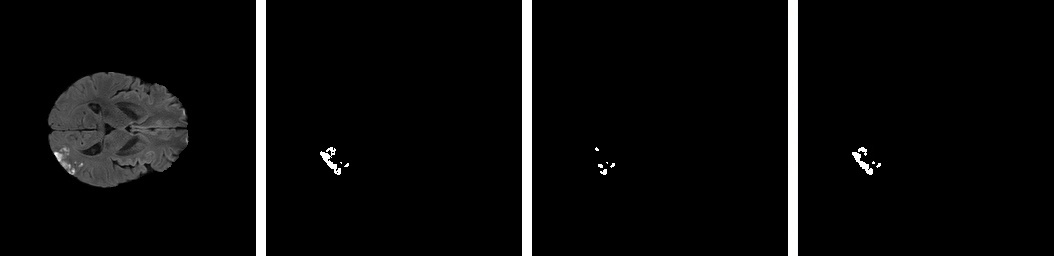}
        \vskip 0.08cm
        \includegraphics[width=\linewidth]{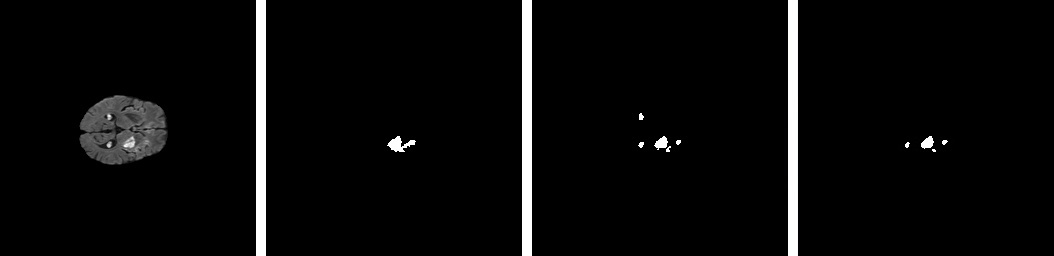}
    \end{minipage}
    \hfill
    \begin{minipage}{0.32\linewidth}
        \centering
        \includegraphics[width=\linewidth]{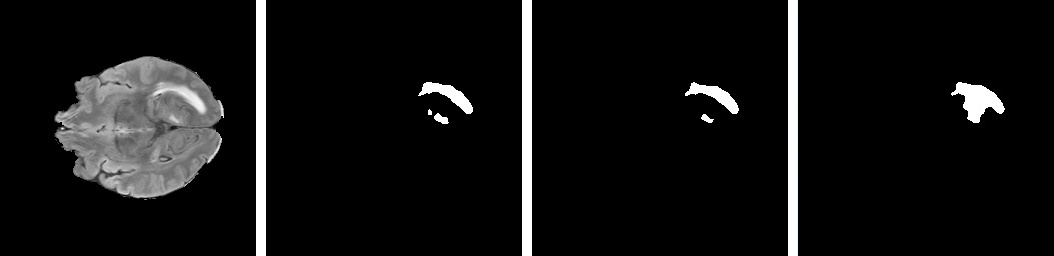}
        \vskip 0.08cm
        \includegraphics[width=\linewidth]{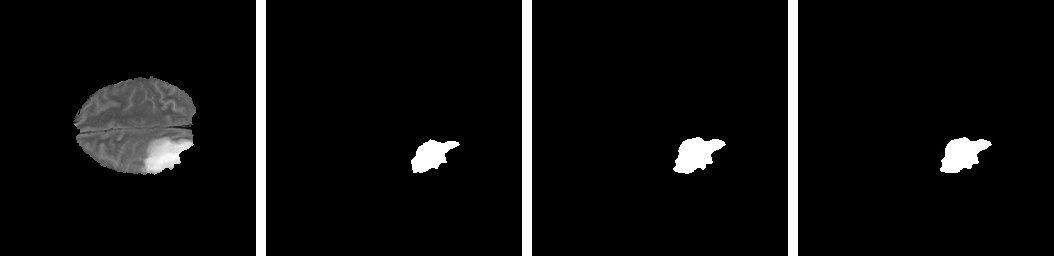}
    \end{minipage}
    \vskip 0.30cm
    \begin{minipage}{0.32\linewidth}
        \centering
        \includegraphics[width=\linewidth]{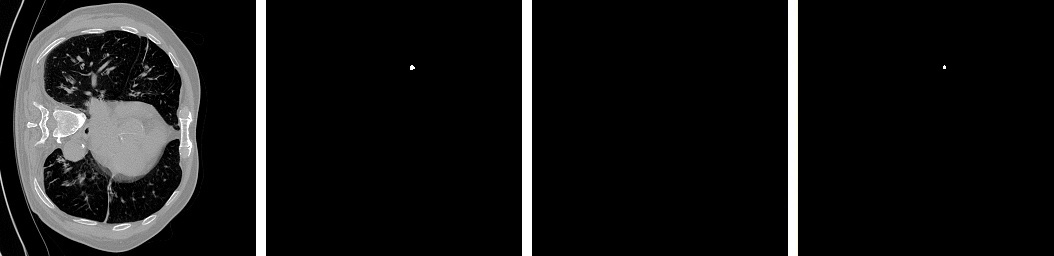}
        \vskip 0.08cm
        \includegraphics[width=\linewidth]{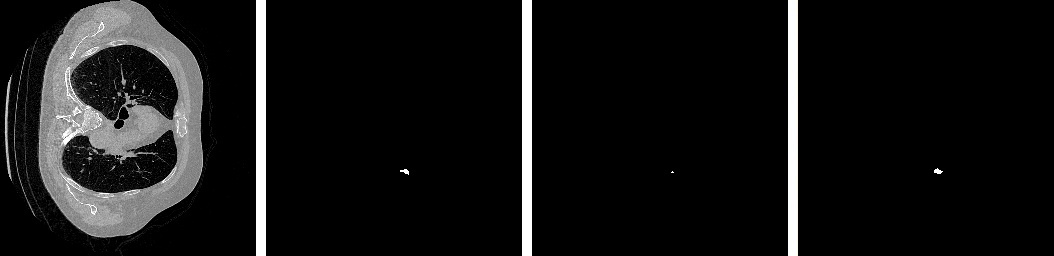}
    \end{minipage}
    \hfill
    \begin{minipage}{0.32\linewidth}
        \centering
        \includegraphics[width=\linewidth]{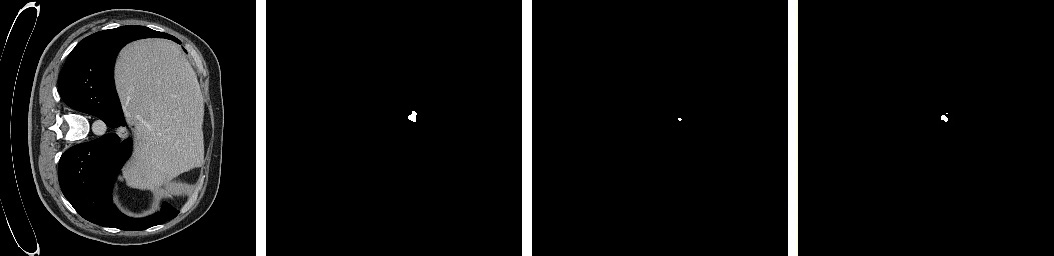}
        \vskip 0.08cm
        \includegraphics[width=\linewidth]{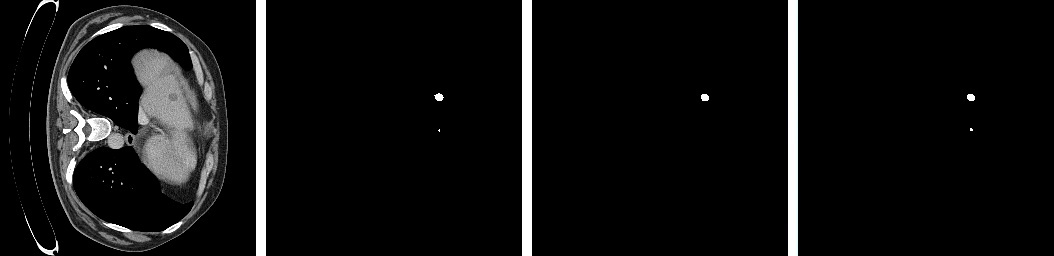}
    \end{minipage}
    \hfill
    \begin{minipage}{0.32\linewidth}
        \centering
        \includegraphics[width=\linewidth]{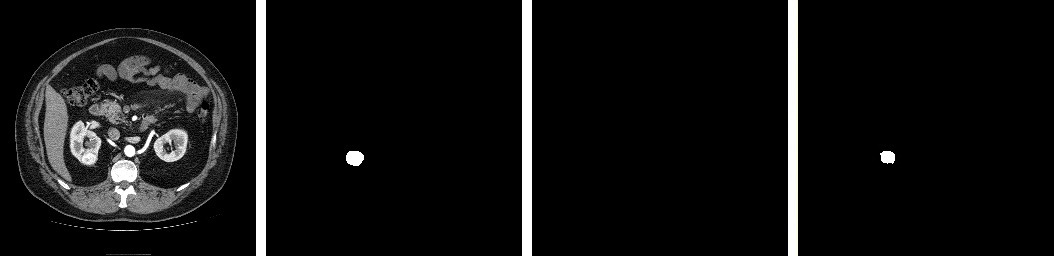}
        \vskip 0.08cm
        \includegraphics[width=\linewidth]{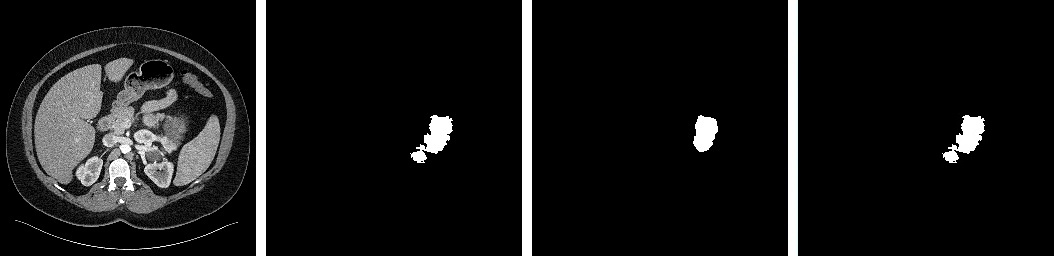}
    \end{minipage}
    \caption{Qualitative results. The top half presents MRI results for BrainMetShare (first column), ISLES (second column), and BraTS (third column), while the bottom half shows CT results for Lung MSD (first column), LiTS (second column), and KiTS (third column). For each dataset, the first row displays TResUNet outputs and the last row shows UNet outputs, including the original image, ground truth, output from the dataset-specific baseline model trained from scratch, and the corresponding student model output distilled from a teacher with the same architecture.}
    \label{fig:combined_results}
\end{figure*}

\medskip
\noindent{\bf Qualitative results.} 
Figure~\ref{fig:combined_results} presents representative qualitative examples for the MRI-based datasets (top rows) and the CT-based datasets (bottom rows), comparing the predictions of dataset-specific baselines and the proposed student models against the ground-truth annotations.

For MRI datasets, the student models generally produce cleaner and more complete lesion segmentations, particularly for small or weakly contrasted regions that are often missed or fragmented by the baseline models. This is especially evident in cases with multiple small metastases or irregular ischemic lesions, where the distilled multi-dataset knowledge helps recover subtle structures while reducing spurious false positives. Boundary adherence is also improved in several examples, in line with the lower HD95 values reported in the quantitative evaluation.

For CT datasets, the qualitative gains are even more pronounced. The student models yield sharper and more anatomically consistent boundaries for lung nodules, liver tumors, and kidney lesions, which are challenging due to their heterogeneous appearance and low contrast with the surrounding tissue. Compared to dataset-specific baselines, the student predictions exhibit fewer boundary irregularities and reduced over-segmentation, corroborating the substantial reductions in HD95 and improvements in IoU observed across CT-based experiments.

Overall, these visual examples complement the quantitative results by illustrating how the proposed cross-domain distillation framework translates into more accurate and robust segmentations in practice, particularly in challenging scenarios involving small lesions, complex shapes, and low-contrast boundaries.

\subsection{Ablation studies}
\label{sec:ablations}

To better understand the contribution of each design choice in the proposed framework, we conduct a series of ablation studies that analyze the impact of (i) target supervision during teacher training, (ii) the number of source datasets used for knowledge aggregation, and (iii) the depth at which teacher features are extracted for distillation. All ablations are performed in the segmentation setting using representative MRI and CT datasets, and are designed to isolate the effect of each component while keeping the remaining parts of the pipeline unchanged.

\medskip
\noindent\textbf{Ablation on target supervision (r).}
Table~\ref{tab:tresunet-mri-ct-r01} reports results on the MRI-based BrainMetShare dataset and the CT-based Lung MSD dataset using TResUNet as the architecture for all models, under two training regimes controlled by the parameter r. When $r=0$, the framework operates in a fully unsupervised domain-adaptation setting, where no labeled samples from the target dataset are used during teacher training. When $r=1$, labeled target examples are incorporated into the loss, enabling supervised optimization of the segmentation objective. Across both datasets, we observe that enabling supervised target adaptation ($r=1$) consistently leads to improved boundary accuracy, as reflected by lower HD95 values. In contrast, overlap-based metrics (IoU, DSC, Precision, and Recall) remain largely comparable between the two settings. This indicates that incorporating labeled target data during teacher training primarily benefits the geometric alignment of predicted boundaries, while the overall region overlap is already well captured in the unsupervised setting. These results highlight the robustness of the proposed framework, which remains effective even when target annotations are unavailable, while still benefiting from supervision when it can be exploited.


\medskip
\noindent{\bf Ablation on the number of sources.} In an additional experiment, we employed our method using the TResUNet architecture but using just a \emph{single} source dataset. Results in table \ref{tab:brain-lung-ablation} were correlated with the ones presented in Table~\ref{tab:all-tresunet-new} but proportional with the number of sources. In this setting we observed an increase in the student’s performance by approximately $+1$ point IoU and a reduction in HD95 by $\sim 1$\,pixel on average. As Table~\ref{tab:all-tresunet-new} shows, incorporating a \emph{second}, complementary source nearly doubles these benefits, yielding about $+2$ points  IoU and a $\sim 3$\, pixel drop in HD95. These results show that segmentation quality scales favorably with the diversity of source knowledge while adding only a moderate extra training cost.

\medskip
\noindent\textbf{Ablation on teacher feature extraction layers.}
Finally, we analyze the impact of the teacher feature depth used for distillation by systematically varying whether features are extracted from the encoder, the bottleneck, or both. As reported in Table~\ref{tab:tresunet-mri-ct-layers}, relying solely on encoder features yields reasonable performance, while bottleneck features provide stronger improvements across most metrics.
Combining encoder and bottleneck features consistently delivers the best results, with IoU gains of approximately $1$--$2$ points and an average HD95 reduction of about 3 pixels compared to single-level configurations. These results confirm the benefit of multi-level distillation, which captures both fine-grained spatial cues and global contextual information, leading to more accurate and stable segmentations with only a modest increase in computational cost.

\begin{table*}[!htbp]
\scriptsize 
\setlength{\tabcolsep}{3pt} 
\caption{Performance on BrainMetShare and Lung MSD using TResUNet as architecture for all models in $r=0$ and $r=1$ setup.}
\centering
\begin{tabular*}{\hsize}{@{\extracolsep{\fill}}@{}c@{} c@{} c@{} c@{} c@{} c@{} c@{} c@{} c@{} c@{} c@{}}
\toprule
\multirow{2}{*}{\textbf{Model}} & \multicolumn{5}{c}{\textbf{BrainMetShare}} & \multicolumn{5}{c}{\textbf{Lung MSD}} \\
\cmidrule(lr){2-6} \cmidrule(lr){7-11}
& \textbf{IoU} & \textbf{DSC} & \textbf{Recall} & \textbf{Prec} & \textbf{HD95} 
& \textbf{IoU} & \textbf{DSC} & \textbf{Recall} & \textbf{Prec} & \textbf{HD95} \\
\midrule
$\text{Multi-head}$    & 75.33 & 78.15 & 81.74 & 85.31 & 13.24 & 81.22 & 83.16 & 83.99 & 89.21 & 10.21 \\ 
$\text{Dataset-specific}$      & 82.84 & 85.19 & 88.21 & 92.62 & 10.99 & 88.06 & 90.53 & 90.86 & 96.48 & 7.71 \\
$\text{Joint teacher $r=0$}$ & \textbf{83.61} & \textbf{85.97} & \textbf{88.24} & \textbf{93.34} & 10.68  & 88.37 & 90.32 & 91.03 & \textbf{96.28} & 6.98 \\
$\text{Joint teacher $r=1$}$ & 83.52 & 85.89 & 88.16 & 93.22 & \textbf{10.16}  & \textbf{88.56} & \textbf{90.83} & \textbf{91.16} & 95.93 & \textbf{6.78} \\
$\text{Student $r=0$}$ & \textbf{86.22} & \textbf{88.57} & \textbf{89.81} & \textbf{94.75} & 8.23  & 90.07 & 92.57 & 93.37 & \textbf{95.78} & 4.99 \\
$\text{Student $r=1$}$ & 86.18 & 88.55 & 89.76 & 94.73 & \textbf{7.95}  & \textbf{90.09} & \textbf{92.72} & \textbf{93.61} & 95.40 & \textbf{4.23} \\
\bottomrule
\end{tabular*}
\label{tab:tresunet-mri-ct-r01}
\end{table*}

\begin{table*}[!htbp]
\scriptsize
\setlength{\tabcolsep}{3pt}
\caption{Ablation on the number of sources: Performance on BrainMetShare (MRI) and Lung MSD (CT) datasets using TResUNet. Joint teacher and student results are shown for single-source and two-source settings. Single-source performance is slightly lower than two-source models, consistent with the text.}
\centering
\begin{tabular*}{\hsize}{@{\extracolsep{\fill}}c c c c c c c c c c c}
\toprule
\multirow{2}{*}{\textbf{Model}} & \multicolumn{5}{c}{\textbf{BrainMetShare}} & \multicolumn{5}{c}{\textbf{Lung MSD}} \\
\cmidrule(lr){2-6} \cmidrule(lr){7-11}
& \textbf{IoU} & \textbf{DSC} & \textbf{Recall} & \textbf{Prec} & \textbf{HD95} 
& \textbf{IoU} & \textbf{DSC} & \textbf{Recall} & \textbf{Prec} & \textbf{HD95} \\
\midrule
Multi-head         & 75.33 & 78.15 & 81.74 & 85.31 & 13.24 & 81.22 & 83.16 & 83.99 & 89.21 & 10.21 \\
Dataset-specific   & 82.84 & 85.19 & 88.21 & 92.62 & 10.99 & 88.06 & 90.53 & 90.86 & 96.48 & 7.71 \\
Joint teacher (source 1 only) & 82.52 & 85.09 & 87.11 & 91.02 & 11.16 & 87.05 & 89.50 & 90.12 & 95.00 & 8.70 \\
Joint teacher (source 2 only) & 82.61 & 85.17 & 87.34 & 91.15 & 11.12 & 87.15 & 89.63 & 90.21 & 95.05 & 8.65 \\
Joint teacher (both sources)  & 83.52 & 85.89 & 88.16 & 93.22 & 10.16 & 88.56 & 90.83 & 91.16 & \textbf{95.93} & 6.78 \\
Student (source 1 only)       & 85.20 & 87.50 & 88.90 & 93.50 & 8.95  & 88.10 & 90.50 & 91.40 & 94.50 & 5.17 \\
Student (source 2 only)       & 85.30 & 87.62 & 88.95 & 93.62 & 8.90  & 88.25 & 90.65 & 91.50 & 94.60 & 5.10 \\
Student (both sources)        & \textbf{86.18} & \textbf{88.55} & \textbf{89.76} & \textbf{94.73} & \textbf{7.95}  & \textbf{90.09} & \textbf{92.72} & \textbf{93.61} & 95.40 & \textbf{4.23} \\
\bottomrule
\end{tabular*}
\label{tab:brain-lung-ablation}
\end{table*}

\begin{table*}[!htbp]
\scriptsize 
\setlength{\tabcolsep}{3pt} 
\caption{Performance on BrainMetShare and Lung MSD using TResUNet as architecture for all models taking features from encoder, bottleneck or both.}
\centering
\begin{tabular*}{\hsize}{@{\extracolsep{\fill}}@{}c@{} c@{} c@{} c@{} c@{} c@{} c@{} c@{} c@{} c@{} c@{}}
\toprule
\multirow{2}{*}{\textbf{Model}} & \multicolumn{5}{c}{\textbf{BrainMetShare}} & \multicolumn{5}{c}{\textbf{Lung MSD}} \\
\cmidrule(lr){2-6} \cmidrule(lr){7-11}
& \textbf{IoU} & \textbf{DSC} & \textbf{Recall} & \textbf{Prec} & \textbf{HD95} 
& \textbf{IoU} & \textbf{DSC} & \textbf{Recall} & \textbf{Prec} & \textbf{HD95} \\
\midrule
$\text{Joint teacher (encoder)}$              & 74.83 & 74.56 & 74.44 & \textbf{99.97} & 25.17 & 80.31 & 83.54 & 83.78 & \textbf{97.44} & 15.33 \\
$\text{Joint teacher (bottleneck)}$           & 82.75 & 85.46 & 87.04 & 93.89 & 10.77 & 86.22 & 88.95 & 89.33 & 93.97 & 6.93 \\
$\text{Joint teacher (encoder + bottleneck)}$ & \textbf{83.52} & \textbf{85.89} & \textbf{88.16} & 93.22 & \textbf{10.16} & \textbf{88.56} & \textbf{90.83} & \textbf{91.16} & 95.93 & \textbf{6.78} \\
\bottomrule
\end{tabular*}
\label{tab:tresunet-mri-ct-layers}
\end{table*}


\section{Image Classification Evaluation}

In this section, we evaluate the proposed multi-dataset teacher-student framework on image classification tasks. In contrast to segmentation and detection, classification focuses on learning robust global representations that discriminate between disease categories at the image level. We consider two clinically distinct application domains: pulmonary disease classification from chest X-ray images and cerebral degenerative disease classification from brain MRI. We assess whether multi-dataset knowledge distillation improves generalization across heterogeneous data sources within each domain.

\subsection{Datasets}
\noindent \textbf{COVIDx-CXR}~\cite{pavlova2022covidxcxr3largescaleopensource} is a large-scale chest X-ray dataset for COVID-19 detection, comprising 30,386 radiographs from 17,026 patients collected across more than 51 countries. It provides normal, non-COVID pneumonia, and COVID-19 classes with standardized train/validation/test partitions for benchmarking deep learning models. 

\medskip
\noindent \textbf{RT-PCR Covid19}~\cite{cohen2020covid19imagedatacollection} 
is a continually updated public dataset of chest X-ray (CXR) and CT images with hierarchical diagnostic labels for COVID-19 and other pulmonary conditions. It provides CXR subsets in posteroanterior (PA) / anteroposterior (AP) views and AP-supine (AP taken while lying down), for example, 481 PA/AP and 173 AP-supine images in one snapshot, along with rich metadata and annotations such as lung bounding boxes/segmentations and severity scores (e.g., Brixia, a CXR severity score from 0–18).

\medskip
\noindent\textbf{COVID-QU-Ex}~\cite{TAHIR2021105002, tahir2021covidquex, rahman2021imageenhancement, degerli2021infectionmap, chowdhury2020screening} is a large chest X-ray dataset designed for COVID-19 analysis, containing 33,920 images with ground-truth lung segmentation masks for every sample. It includes 11,956 COVID-19 cases, 11,263 non-COVID infections (viral or bacterial pneumonia), and 10,701 normal images. Although originally curated for segmentation and localization, the dataset also supports robust image-level classification experiments with standardized splits.

\medskip
\noindent \textbf{OASIS MRI}~\cite{marcus2007oasis} is a dataset of approximately 80,000 brain MRI images from 461 subjects, labeled into four classes of Alzheimer’s progression using Clinical Dementia Rating (CDR) scale: non-demented, very mild demented, mild demented, and demented. It supports benchmarking for early Alzheimer’s detection with standardized train/validation/test splits.

\medskip

\noindent\textbf{Alzheimer Disease Neuroimaging Initiative (ADNI)}~\cite{mueller2005adni} is a large-scale, multi-center longitudinal study providing standardized MRI, PET, cerebrospinal fluid (CSF), blood biomarkers, and detailed clinical assessments. In this work, we use MRI-derived data to support classification across the Alzheimer’s disease spectrum, including cognitively normal subjects, mild cognitive impairment (MCI), and Alzheimer’s disease (AD).

\medskip
\noindent{\bf Dataset grouping.} Based on the clinical domains, we group the five datasets into two categories: (1) \textit{Pulmonary disease datasets:} COVIDx-CXR, RT-PCR Covid19, COVID-QU-Ex (for COVID-19 and viral pneumonia classification) and (2) \textit{Cerebral degenerative disease datasets:} OASIS MRI, Alzheimer Disease Neuroimaging Initiative (ADNI) (for different stages of dementia severity spectrum classification). Within each group, one dataset is treated as the target domain, while the remaining datasets serve as complementary sources for multi-dataset teacher training.


\subsection{Teacher-student architectures}
Our framework is compatible with a broad class of image classification networks, allowing flexibility in selecting architectures that best match the characteristics of each dataset. In our experiments, we consider two state-of-the-art models widely used in medical and natural image classification, namely MedViT~\cite{manzari2023medvit} and EfficientNet~\cite{tan2019efficientnet}. These architectures represent complementary design paradigms: (i) a CNN-Transformer hybrid; (ii) a purely convolutional model with principled scaling. Both architectures are evaluated as teachers and students within the proposed multi-dataset distillation framework.

\medskip
\noindent\textbf{MedViT}~\cite{manzari2023medvit}. MedViT is a CNN–Transformer hybrid architecture designed as an alternative to purely convolutional networks~\cite{lecun1998gradient} and standard vision transformers~\cite{dosovitskiy21iclr}, combining the locality bias of CNNs with the global modeling capacity of self-attention. To reduce the quadratic complexity of vanilla self-attention while still capturing long-range dependencies across multiple representation subspaces, MedViT replaces the standard attention operation with an efficient convolution-based mechanism. In addition, it explicitly targets robustness by encouraging smoother decision boundaries: shape information in the high-level feature space is augmented through permutation of feature-wise mean and variance within mini-batches, improving resistance to adversarial perturbations. 
Pretrained on ImageNet~\cite{deng2009imagenet} and extensively evaluated on large-scale medical benchmarks
such as MedMNIST-2D~\cite{yang2020medmnist}, MedViT has demonstrated strong generalization and robustness,
making it well suited for medical image classification.

\medskip
\noindent\textbf{EfficientNet}~\cite{tan2019efficientnet}. EfficientNet introduces a principled compound scaling strategy that jointly and uniformly scales network depth, width, and input resolution using a single coefficient, instead of tuning each dimension independently. 
Starting from a baseline architecture obtained via neural architecture search, this approach yields a family
of models that achieve excellent accuracy-efficiency trade-offs. High-capacity variants, such as EfficientNet-B7, attain state-of-the-art ImageNet performance while using significantly fewer parameters and lower computational cost than conventional CNNs. EfficientNet architectures are known to transfer effectively to a wide range of downstream tasks, including medical image classification, making them a strong and well-established baseline within our framework.

\subsection{Baselines}
We evaluate the proposed classification framework against the same baseline families used throughout the paper:
(i) \emph{dataset-specific classifiers}, trained independently on each target dataset, and
(ii) \emph{multi-head multi-dataset classifiers}, which share a common backbone and use dataset-specific heads.
All baselines follow the definitions and training protocol described in Section~\ref{sec:segmentation_baselines}.

\subsection{Experimental setup} 
We evaluate our classification framework on five datasets spanning two clinical domains: pulmonary disease classification from chest X-ray images and Alzheimer’s disease progression classification from brain MRI. The experimental design aims to assess how effectively the proposed multi-dataset distillation strategy improves classification performance across heterogeneous imaging modalities and disease characteristics.

\medskip
\noindent{\bf Dataset splits and evaluation protocol.} 
Across all datasets, we adopt predefined or commonly used train–test splits, with test-to-train ratios ranging from approximately $0.12$ to $0.25$. Most datasets (\textit{RT-PCR Covid19}, \textit{COVID-QU-Ex}, and \textit{OASIS MRI}) follow splits close to $0.25$, providing a balanced trade-off between training data availability and statistically meaningful evaluation. The remaining datasets employ slightly smaller test partitions (\textit{COVIDx-CXR} at $\approx 0.12$ and \textit{ADNI} at $\approx 0.19$), prioritizing larger training sets while still maintaining sufficiently sized test subsets for reliable performance assessment. This variability reflects the intrinsic characteristics and standard usage protocols of each dataset while ensuring consistent and fair evaluation across experiments.

\medskip
\noindent \textbf{Hyperparameter details for the classification task.} 
All teacher networks are trained using dataset-specific configurations summarized in Tables~\ref{tab:best_params_pulmonary} and \ref{tab:best_params_cerebral}. For pulmonary datasets, baseline models use learning rates between $10^{-5}$ and $5 \times 10^{-5}$, while joint teachers are trained with a learning rate of $10^{-4}$ and dataset-dependent weight decay. For cerebral datasets, baseline learning rates are fixed at $10^{-4}$, whereas joint teachers employ smaller learning rates in the range $[10^{-5},\,5 \times 10^{-5}]$, again with dataset-specific weight decay.
Dropout and attention dropout are applied according to the characteristics of each dataset, as detailed in the corresponding tables. Student architectures are selected per clinical domain: $MedViT\_large$ is used for pulmonary datasets, $MedViT\_small$ for OASIS MRI, and EfficientNet-B0 for ADNI. Student learning rates are set to $10^{-4}$ for MedViT-based models and $10^{-5}$ for EfficientNet-B0, with a uniform weight decay of $10^{-2}$.
%
%
The adversarial alignment loss is scheduled following $ \lambda_{\text{adv}} = \frac{2}{1 + e^{-10p}} - 1, $where $p$ denotes the fraction of completed training steps. This schedule gradually emphasizes domain-invariant feature learning as the classification heads stabilize, allowing the student to first focus on supervised discrimination. All models are trained with early stopping based on validation performance to ensure stable convergence across datasets.

\begin{table*}[!htbp]
\scriptsize
\setlength{\tabcolsep}{3pt}
\caption{Optimal configuration parameters for the oulmonary disease datasets using MedViT\_large as the architecture for the joint-teacher and baseline models.}
\centering
\begin{tabular*}{\hsize}{@{\extracolsep{\fill}} l c c c c c c c c}
\toprule
\textbf{Dataset} & \textbf{Dropout} & \textbf{Attn Dropout} & $\mathbf{lr}_{\text{baseline}}$ & $\mathbf{wd}_{\text{baseline}}$ & \textbf{joint\_ch} & \textbf{FA\_blocks} & $\mathbf{lr}_{\text{joint\_teacher}}$ & $\mathbf{wd}_{\text{joint\_teacher}}$ \\ 
\midrule
COVIDx-CXR      & 0   & 0   & 1e-5 & 1e-2 & 256 & 3 & 1e-4 & 0 \\
RT-PCR Covid19  & 0   & 0   & 5e-5 & 1e-2 & 256 & 3 & 1e-4 & 1e-2 \\
COVID-QU-Ex     & 0.3 & 0.2 & 5e-5 & 0.5  & 256 & 3 & 1e-4 & 1e-2 \\
\bottomrule
\end{tabular*}
\label{tab:best_params_pulmonary}
\end{table*}

\medskip
\noindent \textbf{Classification setup}. 
For both dataset groups, we investigate two complementary layer-wise distillation strategies that differ in where teacher knowledge is injected into the student network:


\begin{itemize}
    \item \textbf{Early-$k$ distillation}, where the first $k$ Fusion Attention Blocks of the joint teacher are distilled into the last $k$ Fusion Attention Blocks of the student, encouraging early semantic alignment while preserving the student’s capacity to adapt at deeper layers;
    \item \textbf{Late-$k$ distillation}, where the last $k$ Fusion Attention Blocks of the joint teacher are distilled into the last $k$ Fusion Attention Blocks of the student, directly aligning high-level representations close to the classification head.
\end{itemize}

For the pulmonary disease datasets, all individual teachers and students share the same backbone architecture and are instantiated as MedViT\_large models, ensuring architectural homogeneity across datasets.

In contrast, the cerebral disease setting adopts a heterogeneous teacher ensemble to reflect cross-architecture knowledge transfer. One teacher is based on MedViT\_small and trained on the ADNI dataset, while a second teacher uses an EfficientNet-B0 backbone trained on OASIS MRI. All student models in this setting employ the MedViT\_small architecture. As a reference baseline, we also include a single MedViT\_small model trained exclusively on OASIS MRI, without any multi-dataset or distillation components.

\subsection{Results} 
We evaluate classification performance using \emph{accuracy}, defined as the proportion of correctly classified samples over the entire test set. Accuracy is a standard and interpretable metric for multi-class medical image classification, allowing direct comparison across datasets and model configurations.

\medskip
To analyze the impact of the proposed teacher–student framework, we consider multiple combinations of teacher and student architectures. Whenever heterogeneous teacher ensembles are employed, the student architecture is chosen to match at least one of the teacher backbones, ensuring that performance gains can be attributed to the proposed multi-level distillation mechanism rather than architectural mismatches.

\medskip
{\bf Pulmonary disease datasets.} 
The best-performing configurations on the pulmonary datasets are reported in Table~\ref{tab:best_params_pulmonary}. For each dataset, we first evaluate dataset-specific baseline models trained from scratch. We then systematically explore joint-teacher configurations by varying key design factors, including the curriculum scheduling strategy and the dimensionality of the shared joint representation. This controlled evaluation enables us to isolate the effect of multi-dataset fusion and feature-level distillation on classification accuracy, and to quantify the contribution of each component to the final performance.

All student architectures are based on the MedViT\_large backbone and are trained using a single loss-weighting configuration, with $cf = 1.0$, $\gamma = 0.3$, $\delta = 0.1$, and $\kappa = 0.4$; the student learning rate and weight decay are set to $1\times10^{-4}$ and $1\times10^{-2}$, respectively.

Preliminary results presented in the Table \ref{tab:pulmonary_early_vs_late} on the COVIDx-CXR dataset indicate that the late-$k$ distillation strategy consistently outperforms early-$k$ distillation. This suggests that transferring knowledge at later stages of the network enables more effective feature alignment and improves overall model performance on pulmonary disease classification.

The final results and a comparative analysis of the proposed models are presented in Table~\ref{tab:summary_best_accuracy_pulmonary}. As shown, the joint teacher model consistently achieves the highest accuracy across all three datasets, highlighting the benefits of joint learning across multiple tasks. The student models (L1–L3) outperform both the baseline and multi-head approaches, demonstrating the effectiveness of the distillation strategy. Among the student variants, deeper distillation generally provides marginal performance gains, with differences between L1, L2, and L3 being relatively small. Overall, these results indicate that the proposed distillation framework successfully transfers knowledge from the joint teacher to the student models, leading to improved classification performance across the evaluated pulmonary disease datasets. We further observed that performance generally exhibits a positive correlation with the number of distilled layers.

\begin{table}[!htbp]
\scriptsize
\setlength{\tabcolsep}{3pt}
\caption{Performance on COVIDx-CXR pulmonary disease dataset using the MedViT\_large architecture for all student, joint-teacher, and baseline models, evaluated with early-$k$ and late-$k$ distillation.}
\centering
\begin{tabular*}{\hsize}{@{\extracolsep{\fill}} l c c}
\toprule
\textbf{Model} & \textbf{COVIDx-CXR early-$k$} & \textbf{COVIDx-CXR late-$k$} \\
\midrule
Multi-head                 & 73.44\% & 73.44\% \\ 
Dataset-specific           & 78.54\% & 78.54\% \\
Joint teacher              & 80.13\% & \textbf{83.58}\% \\
Student L1                 & 79.44\% & 81.40\% \\
Student L2                 & 79.75\% & 81.82\% \\
Student L3                 & 79.88\% & \textbf{81.96}\% \\
\bottomrule
\end{tabular*}
\label{tab:pulmonary_early_vs_late}
\end{table}

\begin{table}[!htbp]
\scriptsize
\setlength{\tabcolsep}{3pt}
\caption{Performance on Pulmonary disease (COVIDx-CXR, RT-PCR Covid19, COVID-QU-Ex) datasets using MedViT\_large as the architecture for all student, joint-teacher, and baseline models.}
\centering
\begin{tabular*}{\hsize}{@{\extracolsep{\fill}} l c c c}
\toprule
\textbf{Model} & \textbf{COVIDx-CXR} & \textbf{RT-PCR Covid19} & \textbf{COVID-QU-Ex} \\
\midrule
Multi-head                 & 73.44\% & 90.13\% & 89.73\% \\ 
Dataset-specific           & 78.54\% & 93.82\% & 91.50\% \\
Joint teacher              & \textbf{83.58}\% & \textbf{97.36}\% & \textbf{95.83}\% \\
Student L1         & 81.40\% & \textbf{95.94}\% & 93.00\% \\
Student L2         & 81.82\% & 95.26\% & \textbf{94.14}\% \\
Student L3         & \textbf{81.96}\% & 94.79\% & 93.42\% \\
\bottomrule
\end{tabular*}
\label{tab:summary_best_accuracy_pulmonary}
\end{table}

\medskip
{\bf Cerebral degenerative disease datasets.} The configurations achieving optimal performance are summarized in Table~\ref{tab:best_params_cerebral}. Baseline models were evaluated for comparison, and multiple joint-teacher configurations were systematically investigated by varying the curriculum learning strategy and the dimensionality of the joint representation vector. This analysis enabled an assessment of how these design choices influenced model performance.

\begin{table*}[!htbp]
\scriptsize
\setlength{\tabcolsep}{3pt}
\caption{Optimal configuration parameters for the cerebral degenerative disease datasets using EfficientNet-B0 and MedViT\_small as the architecture for the joint-teacher and baseline models.}
\centering
\begin{tabular*}{\hsize}{@{\extracolsep{\fill}} l c c c c c c c c c}
\toprule
\textbf{Dataset} & \textbf{Model} & \textbf{Dropout} & \textbf{Attn Dropout} & $\mathbf{lr}_{\text{baseline}}$ & $\mathbf{wd}_{\text{baseline}}$ & \textbf{joint\_ch} & \textbf{FA\_blocks} & $\mathbf{lr}_{\text{joint\_teacher}}$ & $\mathbf{wd}_{\text{joint\_teacher}}$ \\ 
\midrule
OASIS MRI    & EfficientNet-B0 & 0.3 & -   & 1e-4 & 1e-2 & 1024 & 3 & 1e-5 & 0.5  \\
ADNI         & MedViT\_small   & 0.3 & 0.2 & 1e-4 & 1e-2 & 1024 & 3 & 5e-5 & 1e-2 \\
\bottomrule
\end{tabular*}
\label{tab:best_params_cerebral}
\end{table*}

The student architectures employ an EfficientNet-B0 backbone for experiments on the OASIS MRI dataset and a MedViT\_small backbone for those conducted on the ADNI dataset. All models are trained using a single loss-weighting strategy with $cf = 0.5$, $\gamma = 0.2$, $\delta = 0.2$, and $\kappa = 0.3$, while the student learning rate and weight decay are fixed at $1\times10^{-5}$ and $1\times10^{-2}$, respectively.

Preliminary results presented in Table \ref{tab:cerebral_early_vs_late} on the ADNI dataset indicate that the late-$k$ distillation strategy consistently outperforms early-$k$ distillation. This suggests that transferring knowledge at later stages of the network enables more effective feature alignment and improves overall model performance in cerebral degenerative disease classification.

Tables \ref{tab:summary_best_accuracy_cerebral_oasis} and \ref{tab:summary_best_accuracy_cerebral_adni} summarize the performance of the distilled student models across the OASIS MRI and ADNI datasets. For OASIS MRI, a consistent improvement in classification accuracy is observed as the distillation depth increases from L1 to L3, indicating that progressively richer knowledge transfer from the teacher model benefits the student. In particular, the L3 distillation strategy achieves the highest accuracy (77.76\%), outperforming both the baseline student model and alternative training paradigms, including the joint teacher and multi-head approaches.

On the ADNI dataset, distillation also leads to notable gains over the baseline, with the L2 configuration yielding the best student performance (91.25\%). However, unlike OASIS MRI, performance slightly degrades at L3, suggesting that deeper distillation may introduce redundant or less transferable representations for this dataset. This behavior highlights the dataset-dependent nature of optimal distillation depth. Overall, the results demonstrate that knowledge distillation substantially enhances student model performance, enabling compact architectures to achieve accuracy comparable to or exceeding that of more complex teacher-based or joint training methods. Moreover, the observed variation between OASIS MRI and ADNI underscores the importance of tailoring the distillation strategy to dataset characteristics.

\begin{figure*}[H]
    \centering
    \includegraphics[height=0.2\textheight]{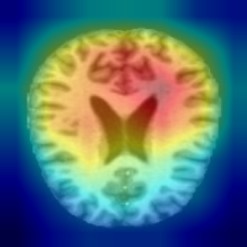}
    \includegraphics[height=0.2\textheight]{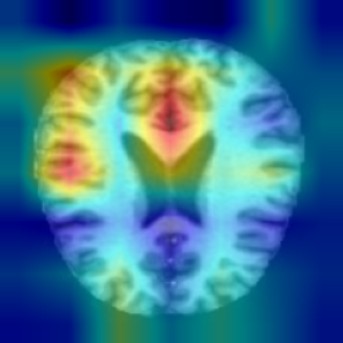}
    \caption{Qualitative results. Attention maps for dataset-specific baseline model trained from scratch (left), and the corresponding student model output distilled from a teacher with the same architecture (right).}
    \label{fig:attention_maps}
\end{figure*}

By qualitatively comparing the attention maps shown in Figure~\ref{fig:attention_maps}, produced by the dataset-specific baseline model trained from scratch and the corresponding student model distilled from a teacher with the same architecture, we observe that the distilled model exhibits more precise and coherent pattern localization, reflecting improved discriminative feature learning.

\begin{table}[!htbp]
\scriptsize
\setlength{\tabcolsep}{3pt}
\caption{Performance on ADNI cerebral degenerative disease dataset using the MedViT\_small architecture for all student, joint-teacher, and baseline models, evaluated with early-$k$ and late-$k$ distillation.}
\centering
\begin{tabular*}{\hsize}{@{\extracolsep{\fill}} l c c}
\toprule
\textbf{Model} & \textbf{ADNI MRI early-$k$} & \textbf{ADNI MRI late-$k$} \\
\midrule
Multi-head         & 84.74\% & 84.74\% \\ 
Dataset-specific   & 88.69\% & 88.69\% \\
Joint teacher      & 89.55\% & \textbf{90.28}\% \\
Student L1         & 89.01\% & 89.94\% \\
Student L2         & 90.21\% & \textbf{91.25}\% \\
Student L3         & 89.94\% & 90.68\% \\
\bottomrule
\end{tabular*}
\label{tab:cerebral_early_vs_late}
\end{table}

\begin{table}[!htbp]
\scriptsize
\caption{Performance on Cerebral degenerative disease OASIS MRI dataset using EfficientNet-B0 as the architecture for joint-teacher and baseline models, and distilling the knowledge into MedViT\_small architecture for the student.}
\centering
\begin{tabular}{l c}
\toprule
\textbf{Method} & \textbf{OASIS MRI} \\
\midrule
Multi-head         & 70.11\%  \\ 
Dataset-specific (MedViT\_small)   & 71.97\%  \\
Dataset-specific (EfficientNet-B0) & 72.29\%  \\
Joint teacher      & \textbf{76.87}\% \\
Student L1         & 74.29\% \\
Student L2         & 76.66\% \\
Student L3         & \textbf{77.76}\% \\
\bottomrule
\end{tabular}
\label{tab:summary_best_accuracy_cerebral_oasis}
\end{table}

\begin{table}[!htbp]
\scriptsize
\setlength{\tabcolsep}{3pt}
\caption{Performance on Cerebral degenerative disease ADNI dataset using MedViT\_small as the architecture for all student, joint-teacher, and baseline models.}
\centering
\begin{tabular}{l c}
\toprule
\textbf{Method} & \textbf{ADNI} \\
\midrule
Multi-head         & 84.74\% \\ 
Dataset-specific   & 88.69\% \\
Joint teacher      & \textbf{90.28}\% \\
Student L1         & 89.94\% \\
Student L2         & \textbf{91.25}\% \\
Student L3         & 90.68\% \\
\bottomrule
\end{tabular}
\label{tab:summary_best_accuracy_cerebral_adni}
\end{table}

\section{Object Detection Evaluation}

In this section, we evaluate the proposed cross-domain teacher–student framework on medical object detection tasks.  Unlike segmentation, detection requires both accurate localization and classification of objects,  posing additional challenges in the presence of domain shifts and heterogeneous data sources.  We follow the same experimental protocol and evaluation philosophy as in the segmentation setting,  adapting only the task-specific components of the pipeline (detection heads and losses), while preserving the domain-adaptation, fusion, and multi-level distillation mechanisms.

\medskip
\noindent\textbf{Lung Cancer CT \& PET-CT}~\cite{lung_cancer_ct_petct} 
is a large-scale dataset comprising 36,631 DICOM images acquired from CT, PET, and fused PET/CT studies of patients with suspected lung cancer. 
The dataset provides bounding-box annotations for lung nodules in PASCAL VOC XML format and is retrospectively collected and stratified by histopathological subtypes, including adenocarcinoma, small-cell, large-cell, and squamous-cell carcinoma.  Its diversity in imaging modalities and pathological presentations makes it well suited for benchmarking lung nodule detection and localization methods.

\medskip
\noindent \textbf{LungCT}~\cite{lungct_roboflow} is a dataset of 2,757 lung CT images curated for computer vision tasks, containing expert-annotated lung nodules. 
Distributed via the Roboflow platform, this dataset supports the development and evaluation of deep learning models for lung nodule detection and localization, particularly in settings with limited training data.

\medskip
\noindent\textbf{DeepLesion}~\cite{yan2018deeplesion} 
is a large and heterogeneous lesion detection dataset consisting of 32,120 axial CT slices from 10,594 imaging studies across 4,427 patients, with a total of 32,735 annotated lesions. 
Lesions span multiple anatomical sites and pathologies, including lung nodules, liver tumors, and enlarged lymph nodes, each annotated with bounding boxes and size measurements. 
In this work, we restrict DeepLesion to images whose lesions are labeled with 
\texttt{Coarse\_lesion\_type} = \textit{lung}, thereby focusing exclusively on thoracic CT slices depicting lung-related abnormalities. 
This filtering yields a lung-specific subset that is directly aligned with our experimental objective of evaluating object detection models on pulmonary lesion localization under realistic clinical variability.

\subsection{Teacher-student architectures}
As in the segmentation and classification settings, the proposed framework is architecture-agnostic and can be instantiated with a wide range of object detection models. This flexibility allows us to select detector families that strike different trade-offs between accuracy, computational cost, and architectural complexity. In our experiments, we focus on two representative and complementary state-of-the-art detectors: Faster R-CNN~\cite{ren2016fasterrcnnrealtimeobject} and RF-DETR~\cite{robinson2025rfdetrneuralarchitecturesearch}. These models exemplify two distinct detection paradigms: (i) a classical two-stage, region-based approach, and (ii) a modern Transformer-based detector. We evaluate both paradigms as teacher and student models within our multi-dataset distillation framework.


\medskip
\noindent\textbf{Faster R-CNN}~\cite{ren2016fasterrcnnrealtimeobject}. Faster R-CNN is a two-stage object detection framework that integrates region proposal generation and object classification into a single, end-to-end trainable network. It introduces a Region Proposal Network (RPN) that shares full-image convolutional features with the detection head, enabling efficient, nearly cost-free generation of class-agnostic region proposals. The RPN densely predicts objectness scores and bounding box offsets at each spatial location, while the second stage refines these proposals and assigns category labels using a Faster R-CNN-style detector. By sharing convolutional features between the RPN and detector, Faster R-CNN substantially reduces computation compared to earlier proposal-based methods, achieving strong accuracy–speed trade-offs and state-of-the-art performance on standard detection benchmarks.

\medskip
\noindent\textbf{RF-DETR}~\cite{robinson2025rfdetrneuralarchitecturesearch}. Robotflow Detection Transformer (RF-DETR) is a lightweight specialist detection transformer designed to adapt open-vocabulary detectors to new domains while offering flexible accuracy-latency trade-offs. RF-DETR builds on DETR, an object detection model that uses a Transformer encoder-decoder architecture instead of the usual anchor boxes and non-maximum suppression used in classical detectors like Faster R-CNN or YOLO. Instead of fine-tuning large vision-language models directly, RF-DETR first fine-tunes a compact DETR-style base network on the target dataset and then applies weight-sharing neural architecture search to evaluate thousands of candidate architectures with different design choices in depth, width, and other structural parameters, all without retraining each configuration from scratch. This procedure yields dataset-specific Pareto-optimal models that transfer effectively to diverse application domains, including real-world settings with out-of-distribution classes. RF-DETR attains state-of-the-art real-time performance on benchmarks such as COCO and Roboflow100-VL, including the first real-time detector to surpass 60 mAP on COCO dataset, while maintaining competitive inference speed.


\subsection{Baselines}

For object detection, we adopt the same baseline families used in the segmentation and classification evaluations, ensuring a consistent experimental protocol across tasks. Specifically, we compare against (i) \emph{dataset-specific detectors}, trained independently on each target dataset, and (ii) \emph{multi-head multi-dataset detectors}, which share a common backbone across datasets with dataset-specific detection heads.
Both Faster R-CNN and RF-DETR are used within these baseline configurations, matching the student architectures. As in previous sections, these baselines allow us to disentangle the effects of isolated training, joint multi-dataset learning, and explicit teacher–student knowledge distillation in the context of object detection.

\subsection{Experimental setup}

We evaluate the proposed object detection framework on three CT-based datasets focused on pulmonary disease. The experimental design aims to assess how effectively multi-dataset teacher–student distillation improves detection accuracy and localization performance across heterogeneous imaging sources and acquisition conditions.

\medskip
\noindent\textbf{Dataset splits and evaluation protocol.}
For all object detection datasets, we adopt a consistent data partitioning strategy with approximately 80\% of the samples used for training and 20\% reserved for testing. This split reflects standard practice in object detection benchmarking, providing sufficient data for model optimization while ensuring statistically meaningful evaluation. Maintaining a uniform train–test ratio across datasets guarantees methodological consistency and enables fair and comparable assessment of detection performance under identical experimental conditions.


\medskip
\noindent \textbf{Hyperparameter details Faster R-CNN and RF-DETR architectures.} All student Faster R-CNN and RF-DETR models are trained with a combination of detection and knowledge distillation (KD) losses. The KD components include batch-wise contrastive loss ($\beta = 0.005$), feature-level alignment ($\gamma = 0.03$), feature-level cosine similarity ($\delta = 0.03$), attention transfer ($\eta = 0.01$), classification/logit KD ($\zeta = 1.0$), and ROI feature imitation ($\rho = 0.10$). These losses are aggregated into a total KD loss and scaled by a curriculum factor $cf = 1$, which can ramp from $0$ to $1$ following a cosine-based schedule over the ramp period. The classification and contrastive components are applied with a temperature of $T = 2.0$, and gradient clipping with maximum norm $5.0$ is used to stabilize training. This curriculum ensures that the student first learns the detection task before progressively aligning with the teacher’s feature and logit representations, with the total KD loss combined with the standard detection loss to guide optimization.

\medskip
%

\noindent\textbf{Detection setup.}
All object detection experiments are conducted using two architectures, Faster R-CNN and RF-DETR, which are treated independently throughout the pipeline. That is, teacher and student models always belong to the same detector family, and no cross-architecture fusion is performed. For a given architecture, we first construct a fusion-based joint teacher by combining multiple dataset-specific teacher detectors, as described in Section~\ref{stage2-joint-teacher}. A single student detector is then trained via knowledge distillation from this joint teacher. The student is optimized using the standard detection loss together with additional feature-level distillation terms, which enforce alignment between the multi-scale backbone (and FPN, when applicable) features of the joint teacher and those of the student. 

\begin{table*}[!htbp]
\scriptsize
\setlength{\tabcolsep}{3pt}
\caption{Performance on Lung CT (DeepLesion, LungPet, and LungCT) datasets using Faster R-CNN as the architecture for all student, joint-teacher, and baseline models.}
\centering
\begin{tabular*}{\hsize}{@{\extracolsep{\fill}}@{}l@{} c@{} c@{} c@{} c@{} c@{} c@{}}
\toprule
\multirow{2}{*}{\textbf{Model}} & \multicolumn{2}{c}{\textbf{DeepLesion}} & \multicolumn{2}{c}{\textbf{LungPet}} & \multicolumn{2}{c}{\textbf{LungCT}} \\
\cmidrule(lr){2-3} \cmidrule(lr){4-5} \cmidrule(lr){6-7}
& \textbf{mAP@[\,.50\,:.95]} & \textbf{mAP@0.50}
& \textbf{mAP@[\,.50\,:.95]} & \textbf{mAP@0.50}
& \textbf{mAP@[\,.50\,:.95]} & \textbf{mAP@0.50} \\
\midrule
Multi-head            & 38.33 & 70.68 & 63.83 & 97.45 & 44.52 & 80.73 \\
Dataset-specific      & 41.20 & 73.91 & \textbf{66.72} & 99.05 & 47.23 & 83.20 \\
Joint teacher         & \textbf{49.94} & \textbf{86.40} & 65.74 & \textbf{99.51} & \textbf{49.31} & \textbf{87.34} \\
Student               & \textbf{45.03} & \textbf{80.46} & 65.61 & 99.28 & \textbf{47.89} & \textbf{85.53} \\
\bottomrule
\end{tabular*}
\label{tab:performance_fasterrcnn}
\end{table*}

\begin{table*}[!htbp]
\scriptsize
\setlength{\tabcolsep}{3pt}
\caption{Performance on Lung CT (DeepLesion, LungPet, and LungCT) datasets using RF-DETR as the architecture for all student, joint-teacher, and baseline models.}
\centering
\begin{tabular*}{\hsize}{@{\extracolsep{\fill}}@{}l@{} c@{} c@{} c@{} c@{} c@{} c@{}}
\toprule
\multirow{2}{*}{\textbf{Model}} & \multicolumn{2}{c}{\textbf{DeepLesion}} & \multicolumn{2}{c}{\textbf{LungPet}} & \multicolumn{2}{c}{\textbf{LungCT}} \\
\cmidrule(lr){2-3} \cmidrule(lr){4-5} \cmidrule(lr){6-7}
& \textbf{mAP@[\,.50\,:.95]} & \textbf{mAP@0.50}
& \textbf{mAP@[\,.50\,:.95]} & \textbf{mAP@0.50}
& \textbf{mAP@[\,.50\,:.95]} & \textbf{mAP@0.50} \\
\midrule
Multi-head     & 47.72 & 78.54 & 70.26 & 96.15 & 61.88 & 90.17 \\
Baseline       & 50.89 & 80.13 & \textbf{73.45} & \textbf{98.96} & 64.69 & 93.84 \\
Joint teacher  & \textbf{52.78} & \textbf{83.95} & 72.89 & 98.74 & \textbf{66.78} & \textbf{94.22} \\
Student        & \textbf{51.84} & \textbf{82.46} & 72.71 & 98.72 & \textbf{65.83} & \textbf{94.04} \\
\bottomrule
\end{tabular*}
\label{tab:performance_rfdetr}
\end{table*}

\subsection{Results}
We evaluate object detection performance using two standard metrics: mean Average Precision at an IoU threshold of 0.5 (mAP@0.50) and mean Average Precision averaged over IoU thresholds from 0.50 to 0.95 (mAP@0.50--0.95). While mAP@0.50 primarily reflects coarse object localization and detection sensitivity, mAP@0.50--0.95 provides a more stringent assessment of localization accuracy across varying overlap thresholds, making it particularly informative for medical detection tasks where precise boundary localization is critical.

Tables~\ref{tab:performance_fasterrcnn} and~\ref{tab:performance_rfdetr} report detection results on the DeepLesion, LungPet, and LungCT datasets using Faster R-CNN and RF-DETR architectures, respectively. Across both detector families, we compare dataset-specific baselines, multi-head multi-dataset models, joint teachers, and distilled student detectors.


\medskip
\noindent\textbf{Faster R-CNN results.}
Using Faster R-CNN, the joint teacher consistently achieves the highest detection performance across all three datasets, particularly on the more challenging DeepLesion benchmark. Compared to the dataset-specific baseline, the joint teacher improves mAP@0.50--0.95 by a substantial margin on DeepLesion (+8.7 points) and yields notable gains on LungCT (+2.1 points), demonstrating the benefit of aggregating complementary representations from multiple datasets. The distilled student detector also outperforms the dataset-specific baseline on DeepLesion (+3.8 mAP@0.50--0.95) and LungCT (+0.6), while maintaining competitive performance on LungPet. Although the student does not fully match the joint teacher in this setting, it retains a large fraction of the teacher’s gains while offering a more compact and deployment-friendly model. The multi-head baseline performs consistently worse, highlighting the limitations of naive multi-dataset training without structured knowledge transfer.

\begin{figure*}[ht!]
    \centering
    \begin{minipage}{0.82\linewidth}
        \centering
        \includegraphics[width=\linewidth]{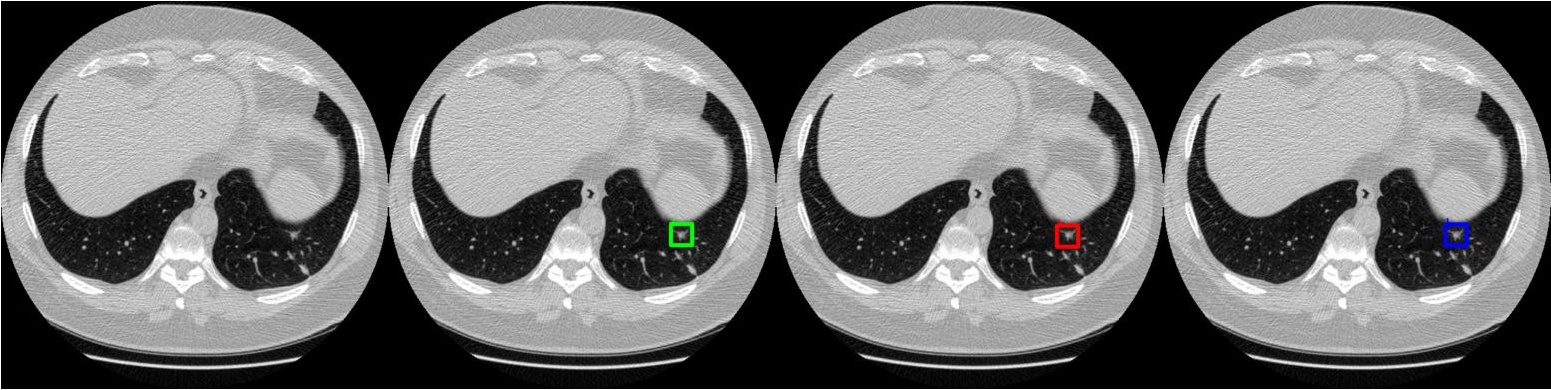}
        \vskip 0.08cm
        \includegraphics[width=\linewidth]{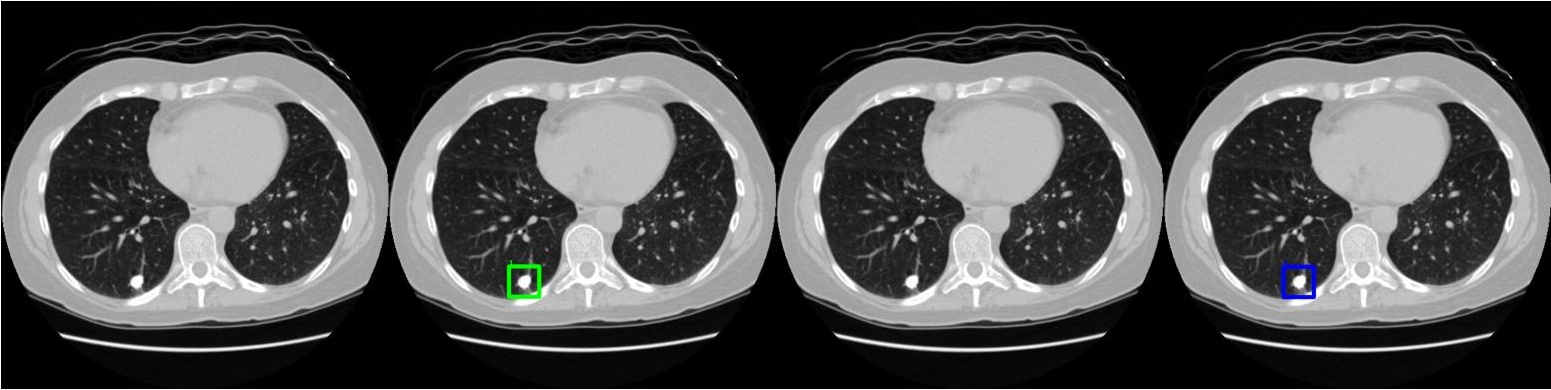}
        \vskip 0.08cm
        \includegraphics[width=\linewidth]{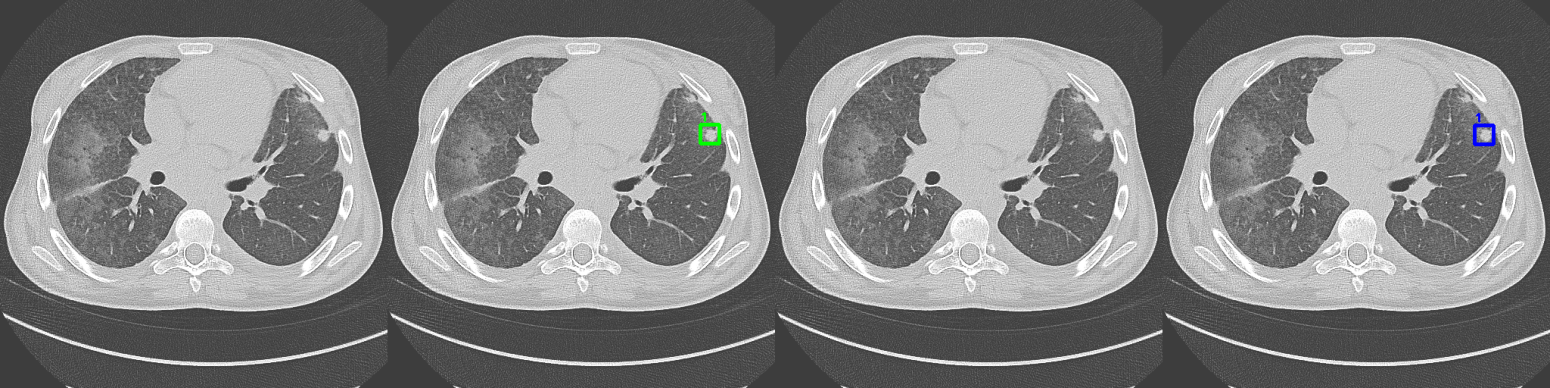}
    \end{minipage}
    \vskip 0.30cm
    \begin{minipage}{0.82\linewidth}
        \centering
        \includegraphics[width=\linewidth]{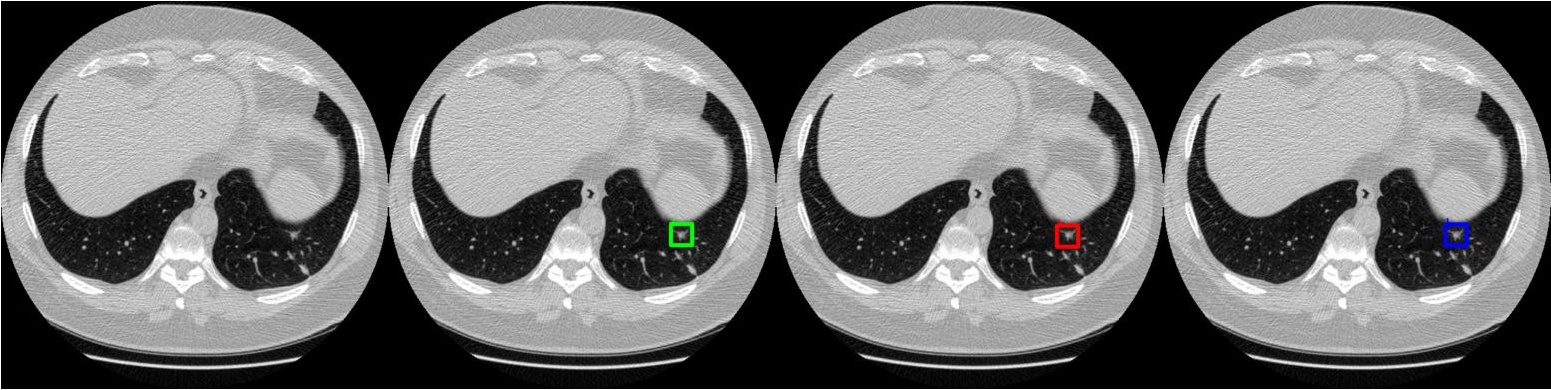}
        \vskip 0.08cm
        \includegraphics[width=\linewidth]{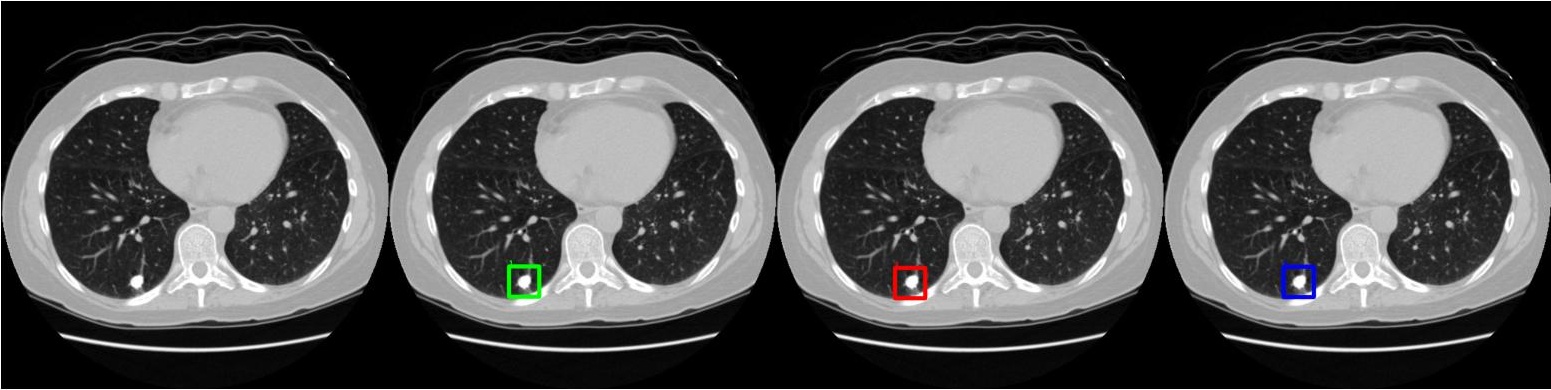}
        \vskip 0.08cm
        \includegraphics[width=\linewidth]{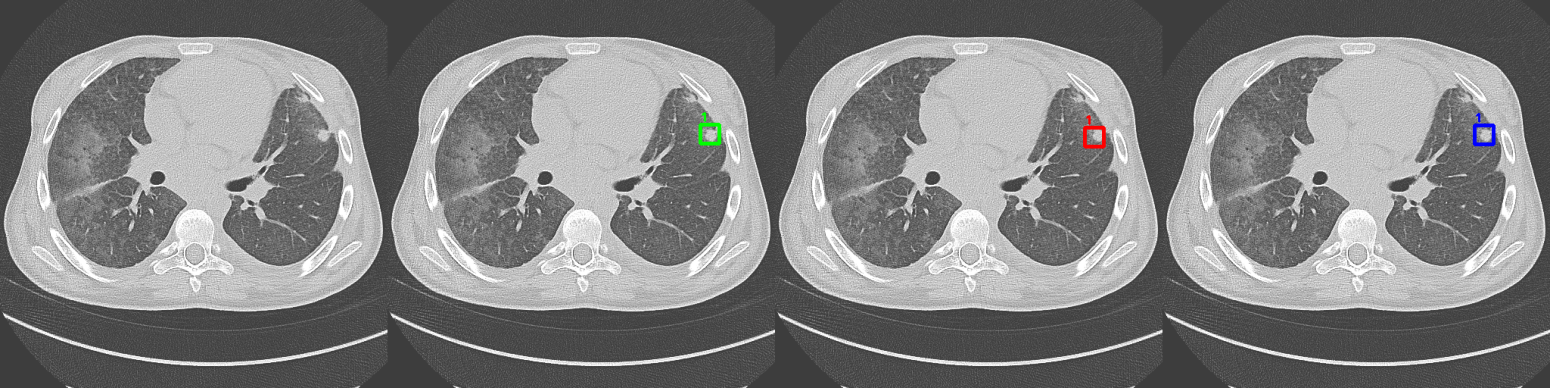}
    \end{minipage}
    \caption{Qualitative object detection results on lung CT datasets. The top half shows Faster R-CNN predictions, while the bottom half shows RF-DETR predictions. For each detector, results are displayed for Lung Cancer CT \& PET-CT (first row), LungCT (second row), and DeepLesion (third row). Within each row, we show (from left to right): the original input image, the ground-truth bounding boxes, the output of the dataset-specific baseline model trained from scratch, and the output of the corresponding student model obtained via multi-dataset knowledge distillation. Best viewed in color. 
    }
    \label{fig:od-combined_results}
\end{figure*}

\begin{figure*}
    \centering
    \includegraphics[width=1\linewidth]{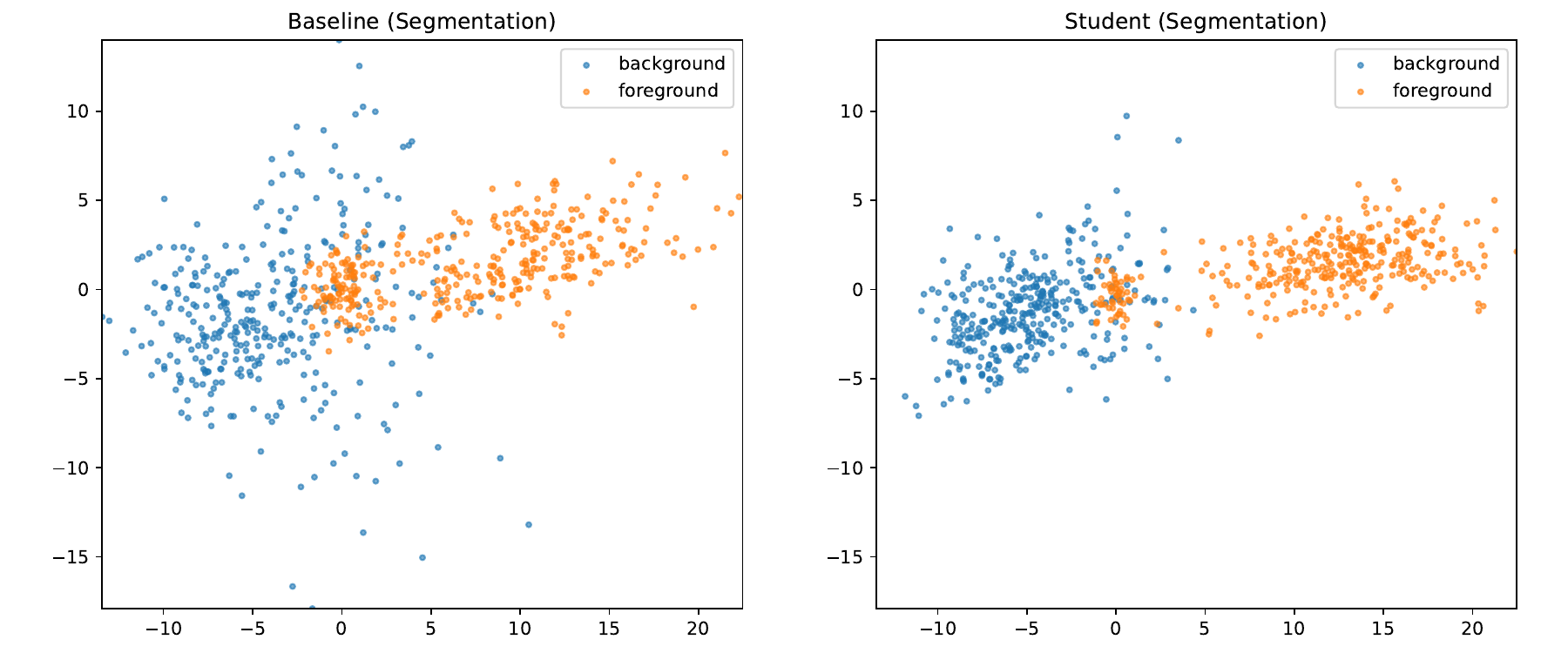}
    \includegraphics[width=1\linewidth]{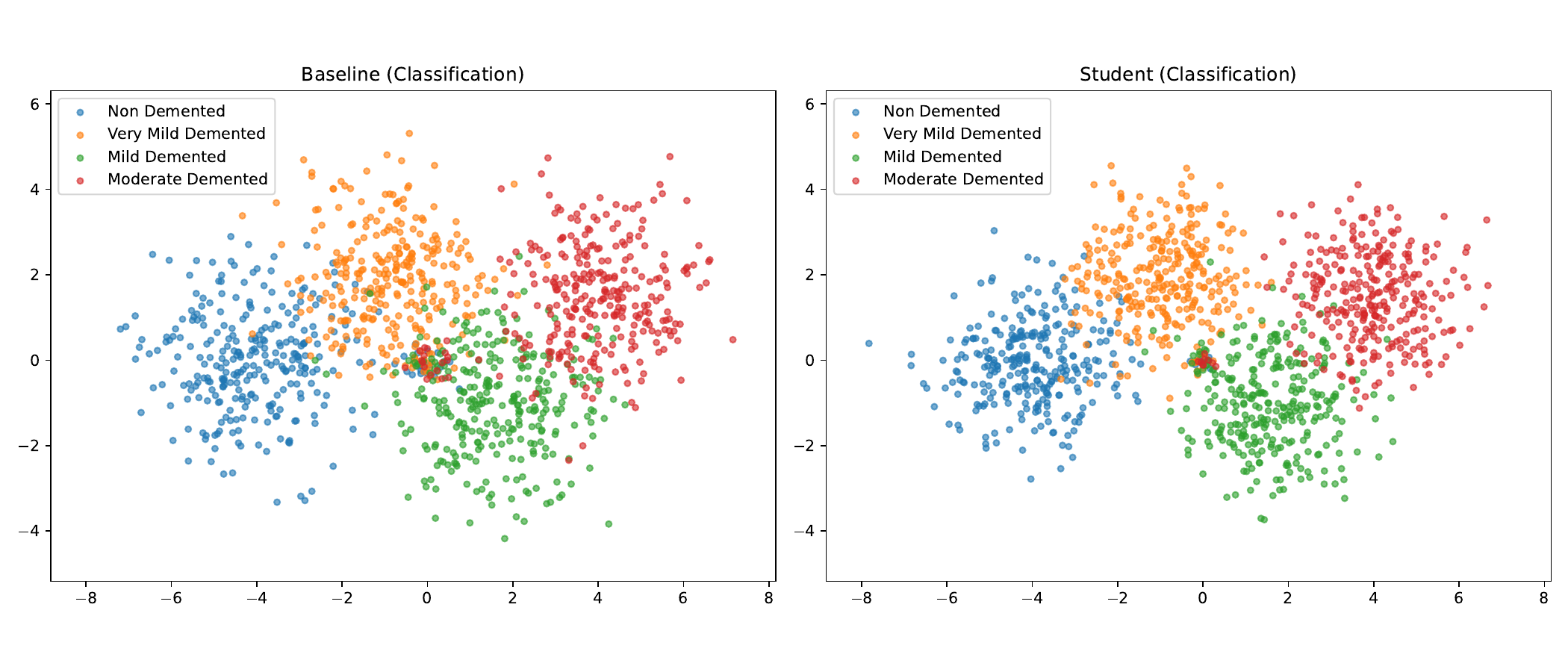}
    \includegraphics[width=1\linewidth]{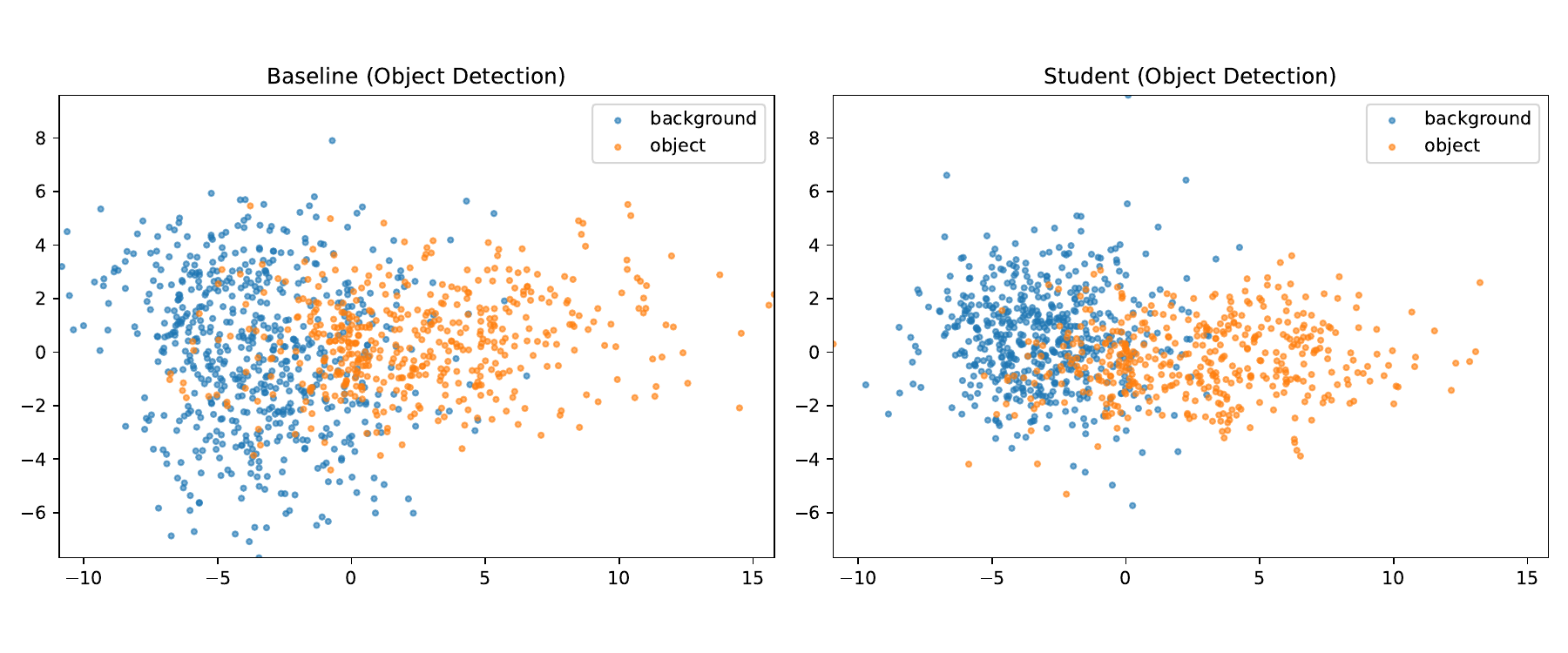}
    \caption{t-SNE visualizations of learned feature representations across tasks. 
    \textbf{Top:} Pixel-level embeddings extracted from the TResUNet bottleneck on the BrainMetShare dataset, showing separation between brain metastases (foreground) and background tissue.
    \textbf{Middle:} Image-level embeddings from the penultimate layer of MedViT on the OASIS MRI dataset, illustrating the clustering of the four diagnostic classes.
    \textbf{Bottom:} Region-level embeddings from RF-DETR decoder queries on the DeepLesion dataset, highlighting the separation between lung lesions and background regions.
    Across all tasks, distilled student models form more compact and better-separated clusters than dataset-specific baselines, indicating improved discriminative representations.}
    \label{fig:placeholder}
\end{figure*}

\medskip
\noindent\textbf{RF-DETR results.}
A similar trend is observed with RF-DETR, where both joint teacher and student models consistently outperform the dataset-specific baseline across all datasets and metrics. On DeepLesion, the joint teacher improves mAP@0.50--0.95 by nearly +2 points over the baseline, while the student recovers most of this gain (+1.0 point). On LungCT, joint teacher and student models achieve the strongest performance, with improvements of approximately +2.1 and +1.1 mAP@0.50--0.95, respectively. Compared to Faster R-CNN, RF-DETR exhibits stronger absolute performance across all datasets, particularly in terms of mAP@0.50--0.95, reflecting its ability to better capture global context and long-range dependencies. Importantly, the proposed distillation framework remains effective across both detector paradigms, indicating that the benefits of multi-dataset fusion and teacher-student transfer are architecture-agnostic.

\medskip
\noindent\textbf{Overall analysis.}
Across both detection architectures and all datasets, the joint teacher consistently delivers the strongest performance, confirming that multi-dataset feature fusion yields richer and more transferable detection representations. The student models, while slightly trailing the joint teacher, consistently outperform dataset-specific and multi-head baselines, demonstrating that the proposed distillation strategy successfully compresses multi-source knowledge into a single detector without significant loss in accuracy. These results validate the effectiveness of the proposed framework for cross-domain object detection in medical imaging, particularly in challenging CT-based lung lesion detection scenarios.

\subsection{Qualitative results}

Figure~\ref{fig:od-combined_results} presents qualitative detection examples on lung CT datasets for both Faster R-CNN (top) and RF-DETR (bottom). Across all datasets, the student detectors distilled from multi-dataset joint teachers produce more accurate and consistent predictions
than dataset-specific baselines trained from scratch.

\subsection{t-SNE visualization}
To further analyze the representational properties induced by the proposed multi-dataset distillation framework, we visualize the learned feature spaces using t-SNE projections.
Figure~\ref{fig:placeholder} presents t-SNE embeddings for segmentation, classification, and object detection, comparing dataset-specific baselines with distilled student models.
Across all three tasks, the visualizations provide qualitative evidence that the proposed framework learns more discriminative and better-structured feature representations.

\medskip
\noindent \textbf{Segmentation (TResUnet on BrainMetShare dataset).} 
For the segmentation task, we analyze pixel-level feature embeddings extracted from the bottleneck layer of TResUNet, which encodes high-level semantic information while retaining spatial context. Feature vectors are sampled from foreground (brain metastases) and background regions, using the ground-truth masks to assign labels. The t-SNE projection reveals that the distilled student model produces a markedly clearer separation between lesion and background embeddings compared to the dataset-specific baseline. Foreground features form more compact and coherent clusters, while background features are pushed further away, indicating improved discriminative power. Residual overlap between clusters is primarily observed near lesion boundaries, reflecting inherent ambiguity in regions with mixed tissue characteristics.

\medskip
\noindent \textbf{Classification (MedViT on the OASIS MRI dataset).} 
In the classification setting, we extract global image-level embeddings from the penultimate layer of MedViT, immediately before the classification head. Each point in the t-SNE plot corresponds to a single MRI scan. The distilled student model exhibits improved class separation and tighter intra-class clustering compared to the baseline, indicating that multi-dataset distillation encourages more structured and class-consistent representations. Partial overlap between certain classes persists, which is expected given the shared anatomical patterns and subtle inter-class differences present in neuro-degenerative imaging.

\medskip
\noindent \textbf{Object Detection (RF-DETR on the Deep Lesion dataset).} 
For object detection, we visualize region-level embeddings extracted from RF-DETR decoder queries. Each embedding is labeled as foreground if it is matched to a ground-truth lung lesion, or background otherwise. The t-SNE visualization shows that the distilled student detector achieves a clearer separation between lesion-related and background embeddings than the dataset-specific baseline. Foreground clusters are more compact and better isolated, suggesting improved consistency in how lesion appearances are encoded. Remaining overlap is largely associated with small or low-contrast lesions and visually ambiguous background regions, which are known to be challenging in CT-based detection.

\section{Conclusions and future work}

In this work, we introduced a unified multi-dataset, cross-domain teacher–student framework for medical image analysis, designed to improve generalization across heterogeneous datasets, imaging modalities, and clinical tasks. The proposed approach integrates domain-adapted source teachers, a joint teacher constructed via multi-level cross-attention fusion, and a curriculum-driven multi-level distillation strategy, enabling effective transfer of complementary knowledge to a compact student model. 

We validated the framework extensively on three core medical imaging tasks, segmentation, classification, and object detection, covering diverse modalities including MRI, CT, PET-CT, and X-ray. Across all tasks, the distilled student models consistently outperformed dataset-specific and multi-head baselines, demonstrating improved overlap accuracy, boundary precision, and detection performance. Detailed ablation studies further highlighted the importance of multi-source supervision, supervised target adaptation, and multi-level feature distillation, while qualitative results and t-SNE visualizations confirmed that the proposed framework learns more compact and discriminative representations.

A key strength of the proposed method lies in its task-agnostic design: while instantiated primarily for medical image segmentation, the same training pipeline extends naturally to classification and object detection with minimal task-specific modifications. This makes the framework broadly applicable to real-world clinical scenarios, where annotated data are scarce, domain shifts are common, and models must operate reliably across institutions and imaging protocols.

Future work will explore extending the framework to additional tasks, such as survival analysis and multi-label diagnosis, incorporating self-supervised pretraining within the joint teacher and scaling the approach to federated and privacy-preserving settings. We also plan to investigate more adaptive fusion mechanisms and uncertainty-aware distillation strategies to further enhance robustness in challenging clinical environments.

{\bf Acknowledgment.} This research is supported by the project “Romanian Hub for Artificial Intelligence - HRIA”, Smart Growth, Digitization and Financial Instruments Program, 2021-2027, MySMIS no. 334906.

\bibliographystyle{cas-model2-names}

\bibliography{cas-refs}


\bio{images/bio/ciprian.jpg}
Ciprian-Mihai Ceausescu earned his B.Sc. in Computers \& Information Technology and M.Sc. in Software Engineering from the University of Bucharest. He currently works as a Teaching Assistant at University of Bucharest and as a researcher within "Romanian Hub for Artificial Intelligence - HRIA". His research interests are Computer Vision and Deep Learning, with applications in the medical field. 
\endbio

\vspace{1em}

\bio{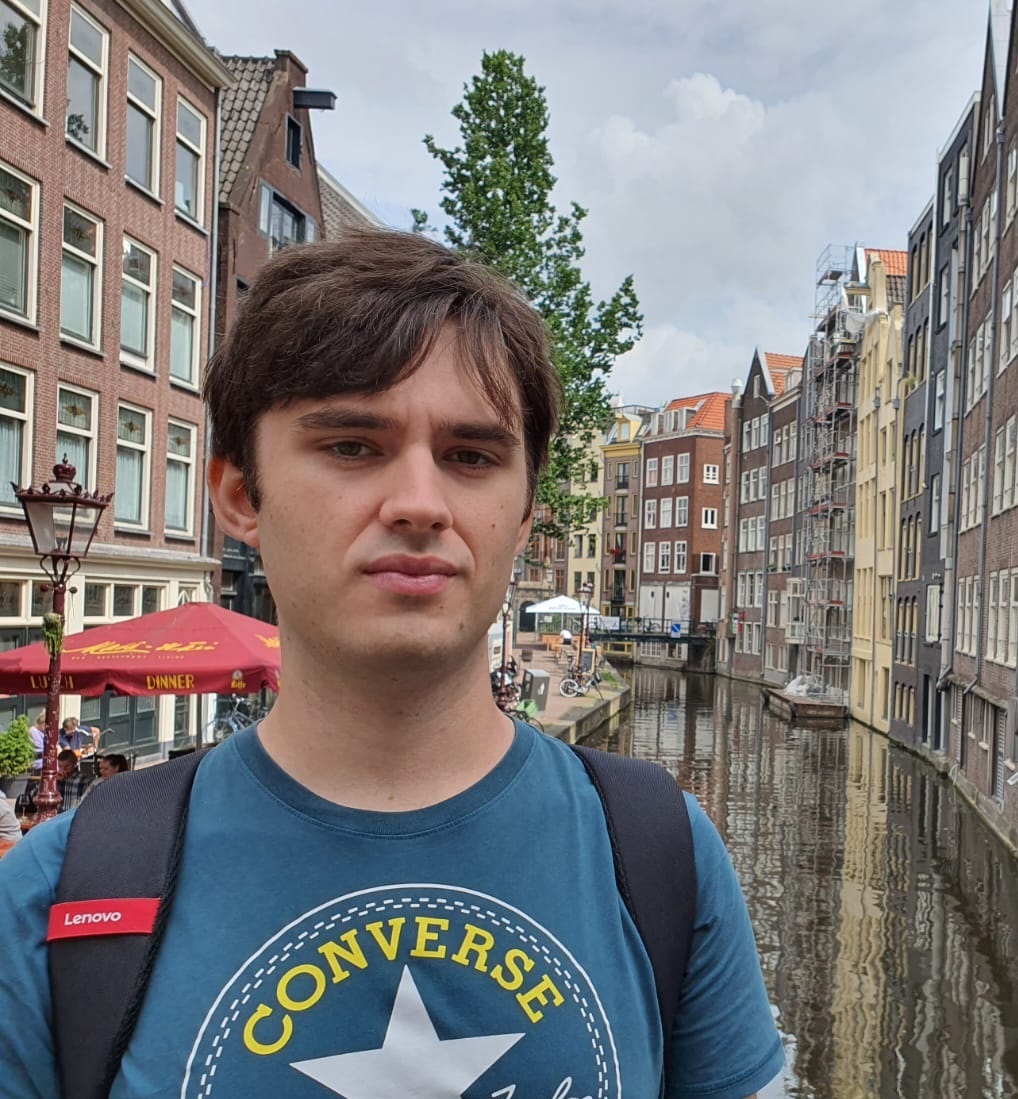}
Ion-Marian Anghelina earned his B.Sc. in Computer Science and M.Sc. in Artificial Intelligence from the University of Bucharest. He currently works as an Applied Scientist at UiPath. His research interests are in Natural Language Processing, with applications in LLMs and Computer Vision.
\endbio

\vspace{4em}

\bio{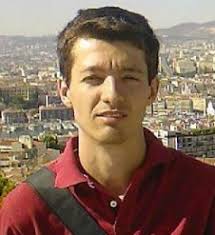}
Dumitru-Bogdan Alexe is an Associate Professor at the Faculty of Mathematics and Computer Science, University of Bucharest, Romania. His research interests include computer vision and machine learning, with emphasis on object detection, weakly supervised learning, video analysis, and medical image understanding. 
\endbio

\end{document}